\documentclass[9pt,twocolumn,twoside]{osajnl}

\usepackage{times}

\usepackage{amsmath}
\usepackage{amssymb}
\usepackage{multirow}
\usepackage{multirow}
\usepackage{mathtools}
\usepackage{units}
\usepackage{subfloat}
\usepackage{float}
\usepackage[]{units}
\usepackage{xcolor}
\usepackage{url}
\usepackage{bbm}
\usepackage{algorithm}
\usepackage{algpseudocode}
\usepackage{pbox}
\usepackage{placeins}
\usepackage{dsfont}

\usepackage[printonlyused]{acronym}
\newacro{AOW}{ANYmal On Wheels}
\newacro{MDP}{Markov Decision Process}
\newacro{RL}{Reinforcement Learning}
\newacro{HRL}{Hierarchical Reinforcement Learning}
\newacro{SRL}{Self-Regulated Learning}
\newacro{PPO}{Proximal Policy Optimization}
\newacro{MPC}{Model Predictive Control}
\newacro{COT}{Cost of Transport}
\newacro{MC}{Minimal Criterion}
\newacro{CPG}{Central Pattern Generator}
\newacro{VAE}{Variational Autoencoder}
\newacro{WFC}{Wave Function Collapse}
\newacro{ICP}{Iterative Closest Point}
\newacro{IMU}{Inertial Measurement Unit}
\newacro{SEA}{Series Elastic Actuator}
\newacro{MLP}{Multi Layer Perceptron}
\newacro{RNN}{Recurrent Neural Network}
\newacro{GRU}{Gated Recurrent Unit}
\newacro{CNN}{Convolutional Neural Network}
\newacro{PDF}{ probability density function}
\newacro{HLC}{high-level controller}
\newacro{LLC}{low-level controller}
\newacro{SPL}{Success weighted by Path Length}
\newacro{LiDAR}{Light Detection and Ranging}

\journal{ol} 

\setboolean{shortarticle}{false}

\title{Learning Robust Autonomous Navigation and Locomotion for Wheeled-Legged Robots}

\author[1,2]{Joonho Lee}
\author[1,3]{Marko Bjelonic}
\author[1,3]{Alexander Reske}
\author[1,3]{Lorenz Wellhausen}
\author[1]{Takahiro Miki}
\author[1]{Marco Hutter}

\affil[1]{Robotic Systems Lab, ETH Zurich, Zurich, Switzerland}
\affil[2]{Neuromeka, Seoul, Korea}
\affil[3]{Swiss-Mile Robotics AG, Zurich, Switzerland}
\affil[2,3]{The work was carried out during their stay at 1}
\affil[*]{Corresponding author: joonho.lee@neuromeka.com}

\dates{This is the accepted version of Science Robotics Vol. 9, Issue 89, adi9641 (2024) \\ DOI: 10.1126/scirobotics.adi9641}

\begin{abstract}
Autonomous wheeled-legged robots have the potential to transform logistics systems, improving operational efficiency and adaptability in urban environments. Navigating urban environments, however, poses unique challenges for robots, necessitating innovative solutions for locomotion and navigation. These challenges include the need for adaptive locomotion across varied terrains and the ability to navigate efficiently around complex dynamic obstacles. This work introduces a fully integrated system comprising adaptive locomotion control, mobility-aware local navigation planning, and large-scale path planning within the city. Using model-free reinforcement learning (RL) techniques and privileged learning, we develop a versatile locomotion controller. This controller achieves efficient and robust locomotion over various rough terrains, facilitated by smooth transitions between walking and driving modes. It is tightly integrated with a learned navigation controller through a hierarchical RL framework, enabling effective navigation through challenging terrain and various obstacles at high speed. Our controllers are integrated into a large-scale urban navigation system and validated by autonomous, kilometer-scale navigation missions conducted in Zurich, Switzerland, and Seville, Spain. These missions demonstrate the system's robustness and adaptability, underscoring the importance of integrated control systems in achieving seamless navigation in complex environments. Our findings support the feasibility of wheeled-legged robots and hierarchical RL for autonomous navigation, with implications for last-mile delivery and beyond.
\end{abstract}

\begin{document}

\maketitle

\section*{Introduction}

\begin{figure*}
    \centering
    \includegraphics[width=\linewidth]{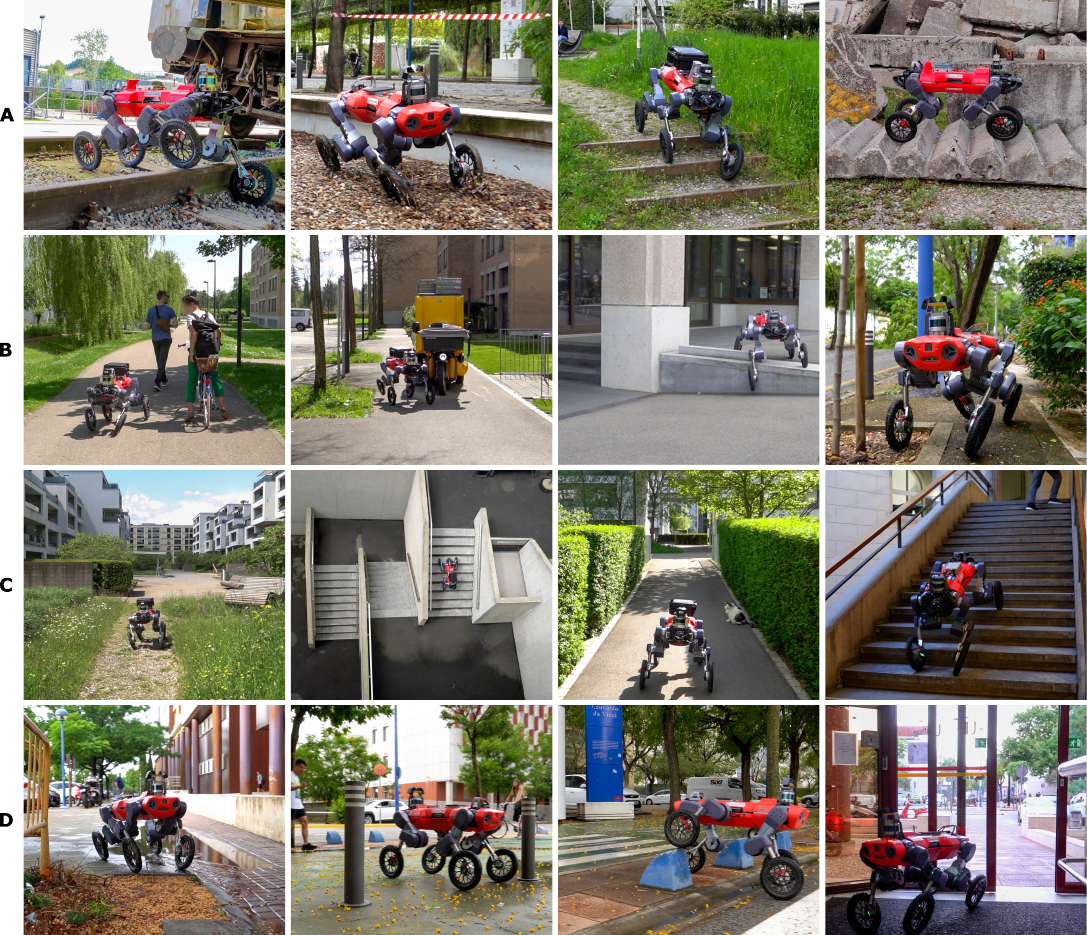}
    \caption{\textbf{Deployments in urban environments}. Our control system for the wheeled-legged robot has undergone extensive validation in various indoor and outdoor locations. The experiments took place in Zurich, Switzerland and in Seville, Spain. \textbf{(A)} Locomotion challenges. \textbf{(B)} Navigation challenges; dynamic and static obstacles, complex terrains, and narrow space. \textbf{(C)} Locations in Zurich. \textbf{(D)} Locations in Seville.}
    \label{fig:highlights}
\end{figure*}

A substantial portion of the population resides in urban areas, leading to a considerable challenge in supply-chain logistics, especially for last-mile deliveries. 
The increasing traffic and demand for faster delivery services put additional pressure on our roads.
By shifting reliance from individual motorized transportation to smart and versatile robotic solutions, we can substantially improve the efficiency of urban delivery. 
Moreover, last-mile delivery routes are not limited to streets but can also include indoor routes, providing an efficient alternative to human labor. 
To fulfill all these roles, robots must be fast and efficient on flat ground while being able to overcome obstacles like stairs. 
Traditional wheeled robots cannot surmount these obstacles effectively, and legged systems alone are inadequate in achieving the necessary velocity and efficiency. 
For instance, the ANYmal robot~\cite{hutter2016anymal} can only operate for a maximum of \unit[1]{hour}\cite{tranzatto2022cerberus, bjelonic2023learning} at half the speed of an average human walking (\unit[2.2]{km/h} on average~\cite{wellhausen2021rough}).

Wheeled-legged robots offer a comprehensive solution that addresses these requirements~\cite{centauro, bjelonic2019keep, klemm2019ascento, Reid2020Mobility}. 
Our research focuses on developing a wheeled-legged robot, as depicted in Figure~\ref{fig:highlights}, where actuated wheels are integrated with its legs~\cite{bjelonic2019keep}. 
Unlike other logistics platforms, this design empowers the robot to operate effectively over long distances, enabling high-speed locomotion on moderate surfaces while maintaining agility on challenging terrains~\cite{bjelonic2022planning,bjelonic2020whole}. 
However, to fully leverage such machines in autonomous real-world applications, it's essential to address several challenges including solving hybrid wheeled-legged locomotion (hybrid locomotion), achieving smooth and efficient navigation, and implementing a complete system that integrates locomotion and navigation modules seamlessly into an autonomous application.

Firstly, hybrid locomotion remains a challenging area in legged robotics. Existing approaches for hybrid locomotion build upon simple heuristics to decide when to step and when to drive~\cite{bjelonic2020whole} or rely on pre-defined gait sequences~\cite{geilinger2018skaterbots, hybrid_minicheetah}.
Most control strategies designed for legged robots incorporate handcrafted gait patterns \cite{bellicoso2018dynamic, jenelten2019dynamic} or motion primitives \cite{lee2020learning, miki2022learning} inspired by nature, but we cannot take observations from biological organisms for wheeled-legged robots.
Determining an effective wheeled-legged gait for each situation is not straightforward, as speed and efficiency heavily depend on the direction of motion and chosen gait. 
For example, minimizing stepping can lead to a lower \ac{COT}~\cite{bjelonic2020whole} for wheeled-legged robots, but traditional methods for legged robots often do not consider gait switching, resulting in sub-optimal outcomes when applied to wheeled-legged robots.
Some directly optimize for \ac{COT}~\cite{xi2016selecting, yang2022fast} and demonstrated improved performance with gait adaptation, but the results are limited to indoor settings or moderate terrains with robots mostly moving forward.
To generate more complex motions that combine driving and walking, trajectory optimization techniques have been utilized to directly optimize gait and discover complex behaviors such as terrain-aware gait and skidding~\cite{bjelonic2022planning, bellegarda2019trajectory}.
However, these methods are computationally expensive and often rely on close-to-optimal initialization.  
Additionally, some of these approaches prioritize computational efficiency at the expense of model accuracy, such as by neglecting the dynamics of wheels, leading to sub-optimal performance on the real robot. 


Secondly, traditional navigation planning methods often overlook the unique characteristics of highly dynamic robots, leading to sub-optimal navigation plans.
Urban environments are mostly covered with flat and open areas that require efficient high-speed traversal to cover large distances. 
At the same time, it is bristled with obstacles like stairs and uneven terrain. 
To achieve speed,  efficiency, and obstacle negotiation capabilities, a navigation algorithm must consider the characteristics of dynamic hybrid locomotion. 
This understanding is crucial for issuing commands that optimize efficiency on flat terrain while maintaining agility when faced with obstacles. 
Many existing approaches \cite{wellhausen2022ART, frey2022locomotion} are based on explicit navigation costs, such as traversability \cite{chavez2018learning, frey2022locomotion}, without considering the robot's whole-body states. 
They focus on generating kinematic navigation plans by sampling-based planning on these estimated cost maps. 
As a result, these approaches often cannot account for various dynamic characteristics of the robot, such as tracking error variations depending on terrain, commanded velocity, or gait. 
Consequently, they may result in frequent turning and stepping actions that can decrease efficiency.

In addition to the previous points, the higher speed capabilities of wheeled-legged robots introduce the need for shorter reaction times, which raises safety concerns and calls for more responsive control systems. 
State-of-the-art sampling-based planners designed for legged robots typically take several seconds to compute a path~\cite{wellhausen2022ART}. 
However, when operating at speeds of multiple meters per second, relying on such planning methods would necessitate long foresight and could result in collisions in dynamic environments.
In dynamic environments or situations involving human presence, ensuring safety requires faster and more frequent decision-making capabilities than what traditional planning methods can provide.


Lastly, attaining autonomy in robotic systems poses a substantial engineering challenge, requiring seamless integration of various sub-modules. 
Traditionally, these sub-modules are developed in isolation with a focus on each component's functionality. 
Their coordination relies heavily on heuristic methods for inter-module communications, and these engineered heuristics often limit smooth and robust operation.
Team Cerberus, for example, undertook the development of an autonomy system for a classic legged robot during the DARPA Subterranean Challenge~\cite{tranzatto2022cerberus}.
The project uncovered substantial insights into operational challenges faced in real-world robotic applications. 
Notably, issues observed during the challenge included robots frequently pausing midway through a path to re-plan or exhibiting zigzag motion while attempting to adhere to a predetermined route.
Such discontinuous or oscillatory behavior can compromise efficiency and hinder the robot's ability to respond to complex dynamic scenarios. 
In some cases, navigation failure may occur when computed navigation paths are not accurately followed~\cite{wellhausen2022ART}.

In this work, we developed a large-scale autonomous navigation system for wheeled-legged robots that enabled seamless coordination between navigation and locomotion controls.
Our approach integrated hybrid locomotion control, which was developed using model-free \ac{RL} and privileged learning~\cite{vapnik2015learning,chen2020learning, lee2020learning}, with a navigation controller optimized through \ac{HRL}.
Both locomotion and navigation controllers were trained using simulated data.
The controllers were integrated into a global navigation framework designed for real-world validations through mock-up delivery missions. Within this framework, digital twins are employed for experiment planning and onboard localization.
Extensive testing in urban areas of Zurich, Switzerland, and Seville, Spain demonstrated the system's autonomy in navigating complex environments, completing kilometer-scale missions across various terrains and obstacles.
Our learned controllers enabled adaptive gait selection, efficient terrain negotiation, and responsive navigation that avoided static and dynamic obstacles safely.
Our practical evaluations underscored the potential of wheeled-legged robots for achieving efficient and robust autonomy in real-world applications.
Additional comparative studies validated the advantage of our tightly integrated navigation controller over traditional systems.

\section*{Results}
\href{https://youtu.be/vJXQG2_85V0}{Movie 1} summarizes our main results.

\begin{figure}
    \centering 
    \includegraphics[width=3.2 in]{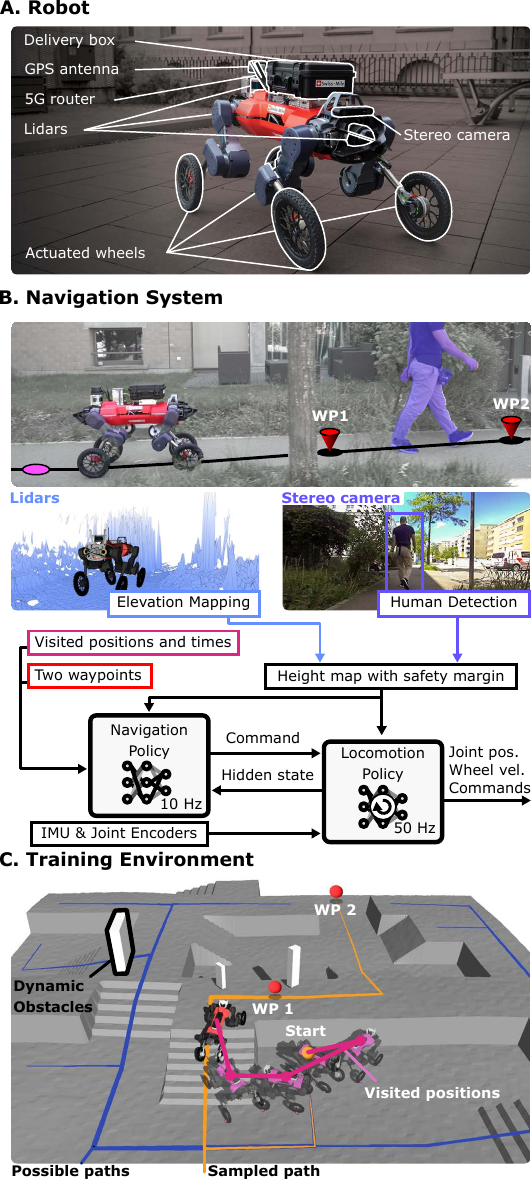}
    \caption{
    \textbf{System overview.} 
    \textbf{(A)} Our wheeled-legged quadrupedal robot is equipped with various payloads for onboard terrain mapping, obstacle detection, and localization.
    \textbf{(B)} Overview of the navigation system. The system is driven by two neural network policies operating at different levels. The high-level navigation policy observes two waypoints (WP 1 and WP 2) and generates target velocity commands for the locomotion policy. The low-level locomotion policy then controls joint actuators and follows the velocity commands. 
    \textbf{(C)} Our training environment is designed to dynamically generate new navigation paths for each episode, optimizing the learning process. 
    By leveraging pre-generated obstacle-free paths, we enhance the navigation capabilities of our system.
    }
    \label{fig:intro}
\end{figure}

\subsection*{System Overview}
We first present a detailed overview of each component comprising our autonomous navigation system. Figure~\ref{fig:intro} provides an overview of our system.

\subsubsection*{Robot}
The wheeled-legged robot used in this work is depicted in Fig.~\ref{fig:intro}A. The robot carried multiple payloads, including three \ac{LiDAR} sensors, an RGB stereo camera at the front, a delivery box, a 5G router, and a GPS antenna. They served various purposes such as localization, terrain mapping, and human detection, contributing to the safety layer.
We integrated an RGB camera with high-frequency object detection capabilities because point cloud-based terrain mapping does not capture dynamic obstacles well.
This allowed for real-time tracking of people within a range of 20 meters. We created a buffer zone in the elevation map by adding an offset around the detected human positions, as detailed in Local Navigation later.
We provide more technical details in Supplementary Materials.

\subsubsection*{Navigation System}
Our navigation system is illustrated in Fig.~\ref{fig:intro}B.
Given a global navigation path, represented by a sequence of graph nodes, we extracted two waypoints, denoted as 'WP1' and 'WP2'. Taking inspiration from the pure-pursuit tracking algorithm~\cite{snider2009pure}, we set two intermediate waypoints by interpolating between the robot's current position projected onto the path and the subsequent graph node, with a fixed look-ahead distance. 

Our robot followed the intermediate waypoints using the \ac{LLC} which is commanded by the \ac{HLC}, both of which are neural networks trained through RL.
\ac{HLC}, fed with the waypoints as input, generated velocity targets for \ac{LLC} at \unit[10]{Hz}, which aligns with the update rate of the onboard elevation mapping~\cite{miki2022elevation}. \ac{LLC}, in turn, generated joint position and wheel velocity commands at \unit[50]{Hz}.

The primary technical contribution of this work lies in the development of our \ac{HLC}. This controller addresses local navigation planning and path-following control together, which traditionally necessitated separate modules.


\subsubsection*{Locomotion controller (LLC)}
We have developed a robust and versatile locomotion controller for wheeled-legged robots by leveraging model-free \ac{RL}.
Our \ac{LLC} is driven by a \ac{RNN}-based policy and builds upon the perceptive locomotion controller by Miki et al.~\cite{miki2022learning}.
We applied modifications to the observation and action space to improve robustness and removed the engineered motion primitives (CPG in \cite{miki2022learning}). Technical details are given in Materials and Methods.

With minimal dependence on human intuition, we achieved a locomotion controller capable of making decisions regarding the gait and transitioning between walking and driving modes.
The locomotion controller is trained in simulation environments thorugh privileged learning~\cite{vapnik2015learning,chen2020learning, lee2020learning}.
During the training, an agent utilizes additional information that is only available during the training phase to enhance the model's performance.
We employed the robot's motion information including velocity and acceleration, terrain properties, and noiseless exteroceptive measurements as privileged information.
During deployment, the final policy only relied on raw measurements from \ac{IMU}, joint encoders, and onboard terrain elevation mapping. 
Similarly to the approach presented by Ji et al.~\cite{ji2022concurrent}, we used the raw \ac{IMU} and encoder measurements instead of using conventional state estimator.
This reduces the use of heuristics for noise filtering and eliminates the need for accurate state estimation for orientation and velocity estimates.
This approach resulted in enhanced robustness when operating on challenging terrains, resulting in fewer failure points in terms of locomotion control.

\subsubsection*{Mobility-aware navigation controller (HLC)}

The \ac{HLC} replaces the traditional navigation setup comprising path planning, path following, and inter-module communication layers~\cite{tranzatto2022cerberus}. 
Instead of explicitly planning future poses and computing reference velocities, our \ac{HLC} directly computes the velocity targets at a high frequency.

The \ac{HLC} processes multiple input modalities including the hidden state of \ac{LLC} policy, terrain height values around the robot, and a sequence of previously visited positions with corresponding visitation times.
Instead of using standard proprioceptive observations, \ac{HLC} accesses the belief state of \ac{LLC}. This latent state captures environmental information such as terrain properties and disturbances, as supported by~\cite{lee2020learning, miki2022learning}.
Additionally, \ac{HLC} processes 20 previously visited positions recorded at \unit[50]{cm} intervals. They span the distance of up to \unit[10]{m}, approximately the usual waypoint spacing.
The history allows \ac{HLC} to make informed decisions based on the robot's prior navigation experience.

Our \ac{HLC} was trained in the simulation environment depicted in Fig.~\ref{fig:intro}C. In every episode, new obstacle-free paths were generated, and two waypoints were sampled along the path at random intervals.

\subsubsection*{Training Environment}
We have adopted the concept of ``Navigation Graph" from computer games~\cite{CryNav, UnrealNav} to provide solvable yet challenging navigation problems to the agent during training (see Fig.~\ref{fig:intro}).
The simulation environment was crafted using a procedural content generation algorithm called \ac{WFC}~\cite{WFC}, and a graph outlining feasible paths and safe areas was constructed with the terrain.
The training environment offered diverse navigation challenges, including detours, dynamic obstacles, rough terrains, and narrow passages. 
As detailed in the Materials and Methods, we combined different obstacles in a controlled manner and rewarded \ac{RL} agents for following the shortest path towards the goal point during training. This approach yielded improved performance compared to policies trained with randomly placed obstacles and goals.

\begin{figure*}
    \centering
    \includegraphics[width=\linewidth]{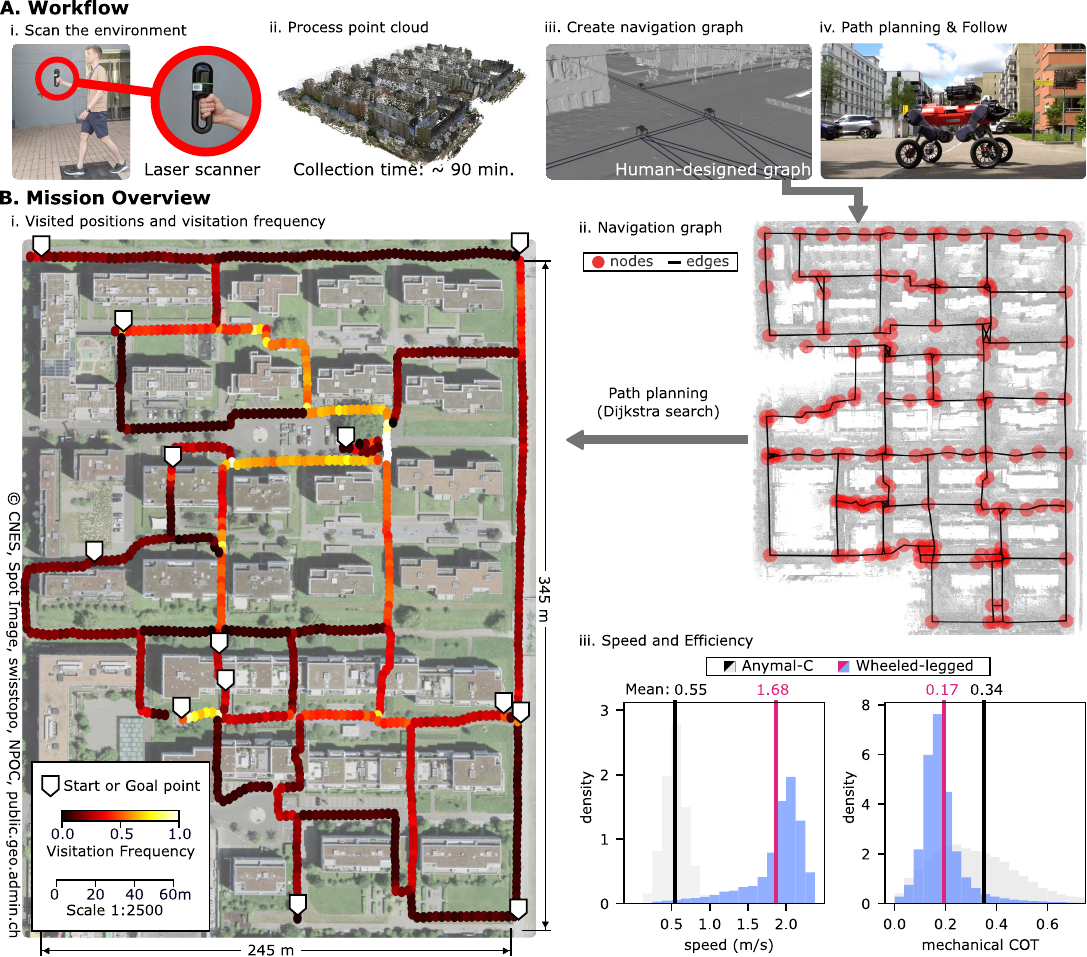}
        \caption{
        \textbf{Large scale autonomous navigation experiment at Glattpark, Zurich.}
\textbf{(A)} Our city navigation workflow begins with offline preparation, involving scanning the test area using a handheld laser scanner and constructing a navigation graph.
\textbf{(B)} The robot autonomously navigated the urban environment to reach 13 predetermined goal points, selected in an arbitrary order. 
\textbf{(B-i, ii)} Path planning within the city was facilitated by the pre-generated navigation graph.
\textbf{(B-iii)} Moving speed and mechanical cost of transport compared to a normal legged robot (ANYmal-C).
}
    \label{fig:glatt}
\end{figure*} 

\subsection*{Kilometer-Scale Autonomous Deployments}

We conducted autonomous navigation missions in different urban environments. These experiments took place in Zurich, Switzerland, and Seville, Spain. The capability of our system is summarized in \href{https://youtu.be/vJXQG2_85V0}{Movie 1}. Additionally, we show one full mission in \href{https://youtu.be/vg_NRWGm270?list=PLE-BQwvVGf8GOIzjIAeuY6qCVgDGTr5jh}{Movie S1} to demonstrate the scale of each experiment.

Fig.~\ref{fig:glatt} summarizes our mock-up delivery mission conducted in Glattpark, Zurich. Our robot covered a total distance of \unit[8.3]{km} with minimal human intervention. 

We first show our workflow in Fig.~\ref{fig:glatt}A. To begin, we employed a handheld laser scanner to capture dense color point clouds of the experimental area. 
The scanning process took approximately 90 minutes to cover a \unit[245]{m} x \unit[345]{m} urban area. 
Subsequently, we georeferenced the point cloud, and the data was converted into a mesh representation, facilitating the creation of a navigation graph and the placement of goal points by a human expert (see Fig.~\ref{fig:glatt}A-iii and Fig.~\ref{fig:glatt}B-ii).
The purpose of the navigation graph is to provide topological guidance and to indicate social preferences, like avoiding landscaping and private property.

During the robot's deployment, it localizes itself with respect to the pre-scanned reference point cloud using its LiDAR, \ac{IMU}, and joint encoder readings. 
With this setup, the robot can be provided with a single GPS goal, and it autonomously navigates toward the target location.  
Selected goal points are sent to the robot via mobile network and the reference path is computed onboard using a shortest-path algorithm~\cite{dijkstra1959note}.
The resulting path is converted into robot-relative coordinates for our navigation policy using LiDAR-localization in the pre-scanned point cloud map.
Note that the point cloud is purely for localization and is not otherwise used for navigation~\cite{jelavic2022open3d}. 
We have found this localization method to be more robust among high-rise buildings than a GPS-based approach.

Fig.~\ref{fig:glatt}B-i illustrates the paths traversed by our robot during multiple long-distance experiments, each lasting more than 30 minutes.
Throughout these experiments, we manually selected 13 distant goal points to maximize coverage of the experimental area. This setup required the robot to navigate diverse obstacles in order to reach each goal point successfully.

Fig.~\ref{fig:glatt}B-iii presents histograms of the speed and mechanical \ac{COT} while the robot was in motion.
We define mechanical \ac{COT} as
\begin{equation}
    COT_{mech} = \sum_{\text{all joints}}{[\tau \dot{\theta}]^{+}}  / (mg|v_{xy}^b|) \quad ,
\end{equation}
where $\tau$ denotes joint torque, $\dot{\theta}$ is joint speed, $mg$ is the total weight, and $|v_{xy}^b|$ is the horizontal speed of the robot's base. This quantity represents positive mechanical power exerted by the actuator per unit weight and unit locomotion speed~\cite{lee2020learning,bjelonic2019keep}.
 
Our robot achieved an average speed of \unit[1.68]{m/s} with a mechanical \ac{COT} of 0.16. In comparison, we provide data on the average speed and \ac{COT} of one ANYmal robot which primarily traversed flat and urban terrains during the DARPA Subterranean challenge~\cite{tranzatto2022cerberus}.
The data is from ANYmal 4 in \cite{anymal_in_darpa}.
Our robot demonstrated three times the speed with a 53\% lower \ac{COT}.
Note that we only compared the output mechanical power. Other major factors contributing to energy loss, such as heat loss and mechanical loss from the actuators' transmission, also come into play during constant walking motions, thus potentially reducing overall efficiency.

The improvement is mainly attributed to the driving mode, which evenly distributes weight across all four legs, keeping leg joints relatively static. Constant stepping led to concentrated loads on fewer legs, requiring higher joint torques and speeds.
During driving, joint actuators contributed almost zero mechanical COT ($\approx0.01$).
Compared to a typical ANYmal robot during locomotion, our robot's wheels exerted approximately 1.2 times the total mechanical power while achieving an average locomotion speed 3.4 times higher.
Upon evaluating the average $\sum \tau^2$ solely for leg joints, our robot exhibited a \unit[16]{\%} lower value, despite being both heavier ($\approx$\unit[12]{kg}) and faster. This quantity is directly related to the heat loss~\cite{hwangbo2019learning,seok2014design}.

\begin{figure}
    \centering
    \includegraphics[width=\columnwidth]{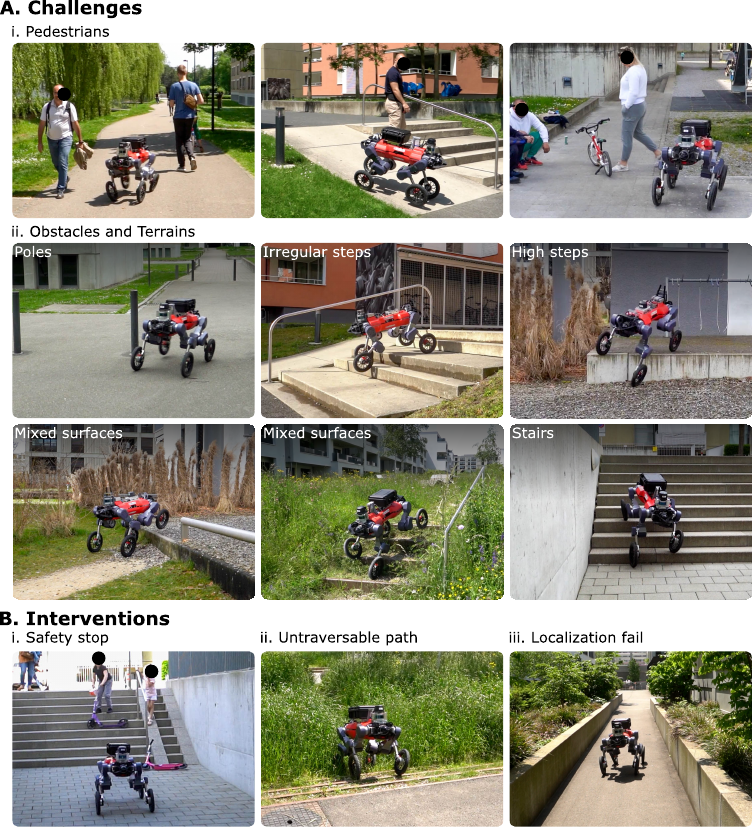}
        \caption{
        \textbf{Challenges in the populated urban environment.}
\textbf{(A)} The urban environment presents various obstacles. Some have to be avoided, such as pedestrians or poles, and others can be traversed, such as stairs or steps. 
    \textbf{(B)} We had to intervene and stop the mission in these three cases.}
    \label{fig:glattB}
\end{figure} 
\begin{figure*}
    \centering
    \includegraphics[width=\linewidth]{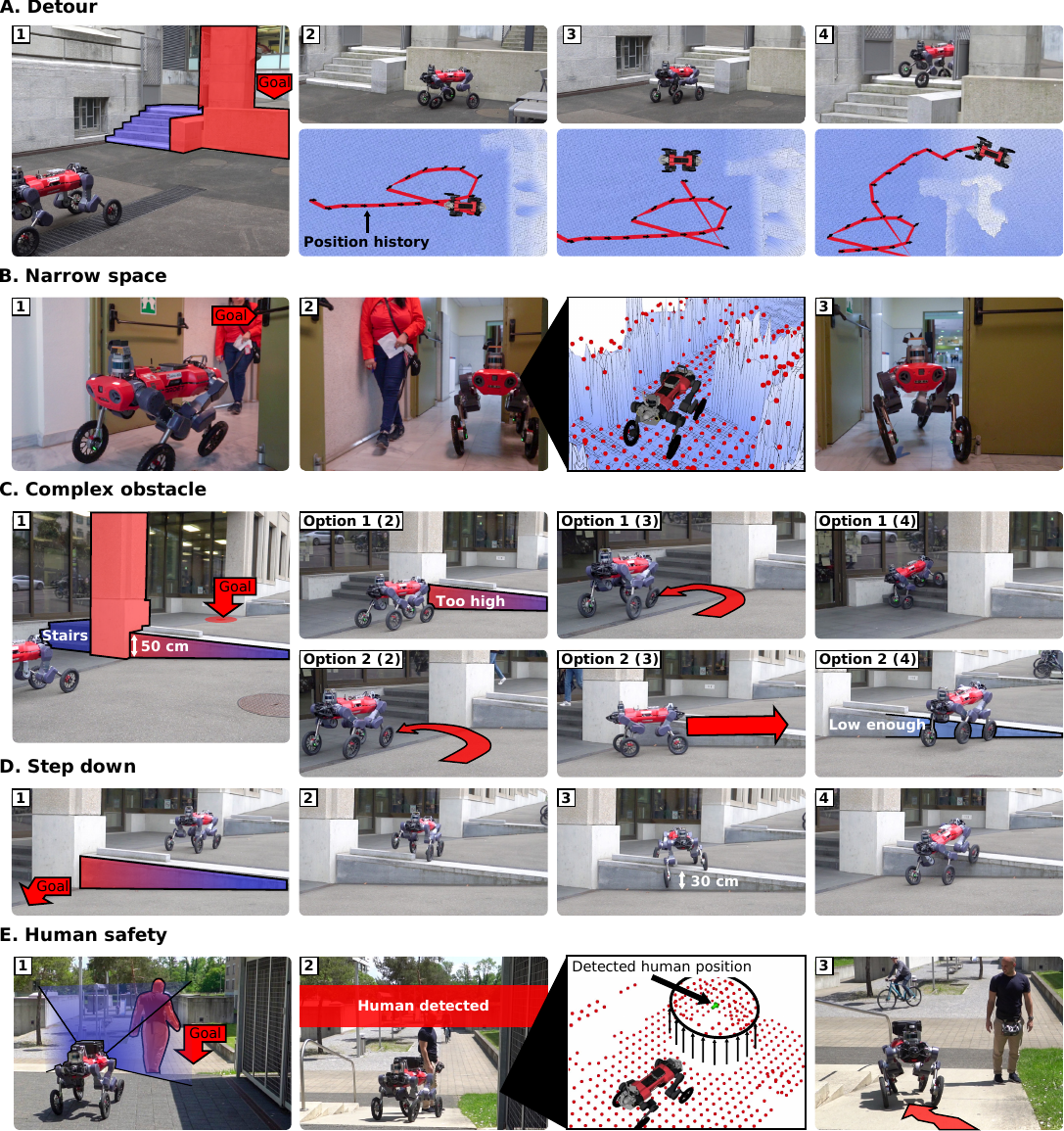}
        \caption{\textbf{Obstacle negotiation}. 
        \textbf{(A)} Our robot navigates around blocked routes by actively exploring the area and finding alternative paths. \textbf{(B)} Safe traversal of a narrow space. \textbf{(C)} Our robot exhibits two different ways to traverse the complex obstacle. \textbf{(C, D)} Our robot shows an asymmetric understanding of traversability, being able to traverse higher steps when going down. \textbf{(D)} We ensure safety around humans by incorporating additional human detection and overriding height scan values. }
        
    \label{fig:obstacles}
\end{figure*} 

Fig.~\ref{fig:glattB} shows the major challenges encountered by our robot, including pedestrians, various obstacles, and non-flat terrains. Our robot demonstrated the capability to navigate around pedestrians in various situations, even on slopes or stairs, as depicted in \href{https://youtu.be/vJXQG2_85V0}{Movie 1} and \href{https://youtu.be/vg_NRWGm270?list=PLE-BQwvVGf8GOIzjIAeuY6qCVgDGTr5jh}{Movie S1}.

Additionally, our robot could avoid thin obstacles, such as the pole shown in the first image of Fig.~\ref{fig:glattB}A-ii, as well as various discrete terrains like steps and stairs.

Because \ac{HLC} and \ac{LLC} are trained to minimize $COT_{mech}$ and $\sum \tau^2$, the robot mostly drove on flat terrain. However, when encountering uneven surfaces, the robot switched to a stepping gait. Importantly, this gait switching is learned without handcrafted heuristics like \ac{CPG} or predefined gait sequences. Furthermore, our controller demonstrated robustness in handling various surfaces, including grass, sand, or gravel, which can be attributed to the privileged training of the \ac{LLC}~\cite{lee2020learning}. 

We intervened during the mission in three circumstances, presented in Fig.~\ref{fig:glattB}B.
Firstly, there were instances where children were in the robot's path.
Although our navigation module would have most likely safely navigated around children, as it did for adults, we prioritized safety and stopped the robot proactively. 

Secondly, we encountered situations where the waypoints were located within untraversable regions. 
For instance, tall grass had grown on a trail between the creation of the navigation graph and the robot deployment, obstructing the path.
Consequently, it presented an obstacle in the local height map used for navigation.
The robot safely stopped in front of the tall grass and we manually triggered global re-planning to go around the obstruction.

Lastly, we encountered challenges with localization in geometrically degenerate environments, such as long corridors.
This meant that the reference path became invalid and provided infeasible, potentially hazardous waypoints.
Our robot's controller was able to operate safely by relying on onboard local terrain mapping but was unable to reach the goal point until localization was recovered.

\subsection*{Local Navigation}

In Fig.~\ref{fig:obstacles}, we present example scenarios that best show the local navigation capability of our system. The sequence of these scenarios can be viewed in \href{https://youtu.be/hMu1CT0Y-Js?list=PLE-BQwvVGf8GOIzjIAeuY6qCVgDGTr5jh}{Movie S2}.

In the first case (Fig.~\ref{fig:obstacles}A), we show the exploratory behavior when the robot encountered a blocked path. The robot reversed and moved along the wall, searching for an opening until it found the stairs leading to the final waypoint. The robot's explicit position memory enabled it to reason about its previously visited positions and navigate through the complex obstacle.

Fig.~\ref{fig:obstacles}B shows our robot's ability to navigate narrow corridors. It safely maneuvered through two doors with a human standing in between, where the gap was as wide as the robot's width. The robot navigated through the narrow space without collision even though human detection was not enabled in this deployment. 
This example showcases the precision and real-time trajectory adjustment of our navigation controller, making it suitable for environments with limited space and tight passageways.

We conducted a test with a complex obstacle depicted in Fig.~\ref{fig:obstacles}C-1. The obstacle consisted of a small staircase on one side and a step with variable height ranging from 0 to \unit[50]{cm} on the other side. When a waypoint was provided above the step, our robot showcased two different approaches.

Initially, when faced with the obstracted route, the robot drove backward and began exploring. During this phase, it could either find the stairs to climb up or continue exploring to find a lower step height. In the second case, after driving along the step, the robot discovered a viable height of approximately \unit[20]{cm}. This example shows the effectiveness of our hierarchical controller, seamlessly adapting its gait based on the terrain, and exhibiting versatility in navigating complex paths.

We observed that \ac{HLC} has an asymmetric understanding of traversability when going up and down the step (Fig.~\ref{fig:obstacles}D). Specifically, the robot was able to traverse higher steps when descending, indicating that it has a more advanced understanding of the terrain compared to cost-map approaches for traversability estimation~\cite{miki2022elevation,frey2022locomotion}. Traditional methods often use symmetric traversability maps that are independent of motion direction, whereas our approach makes decisions based on the current terrain, the robot's state, and the characteristics of the low-level controller.

In Fig.~\ref{fig:obstacles}E, we visualize our strategy for augmenting the static, local elevation map with dynamic obstacles.
We employed camera-based human detection to introduce a height offset within a radius of \unit[50]{cm} around individuals. As the robot encountered a person moving along its path, \ac{HLC}, trained to handle dynamic obstacles of various sizes, maintained a constant distance from the person, enabling the robot to safely overtake.

\subsection*{Hybrid Locomotion}

\begin{figure*}
    \centering
    \includegraphics[width=\linewidth]{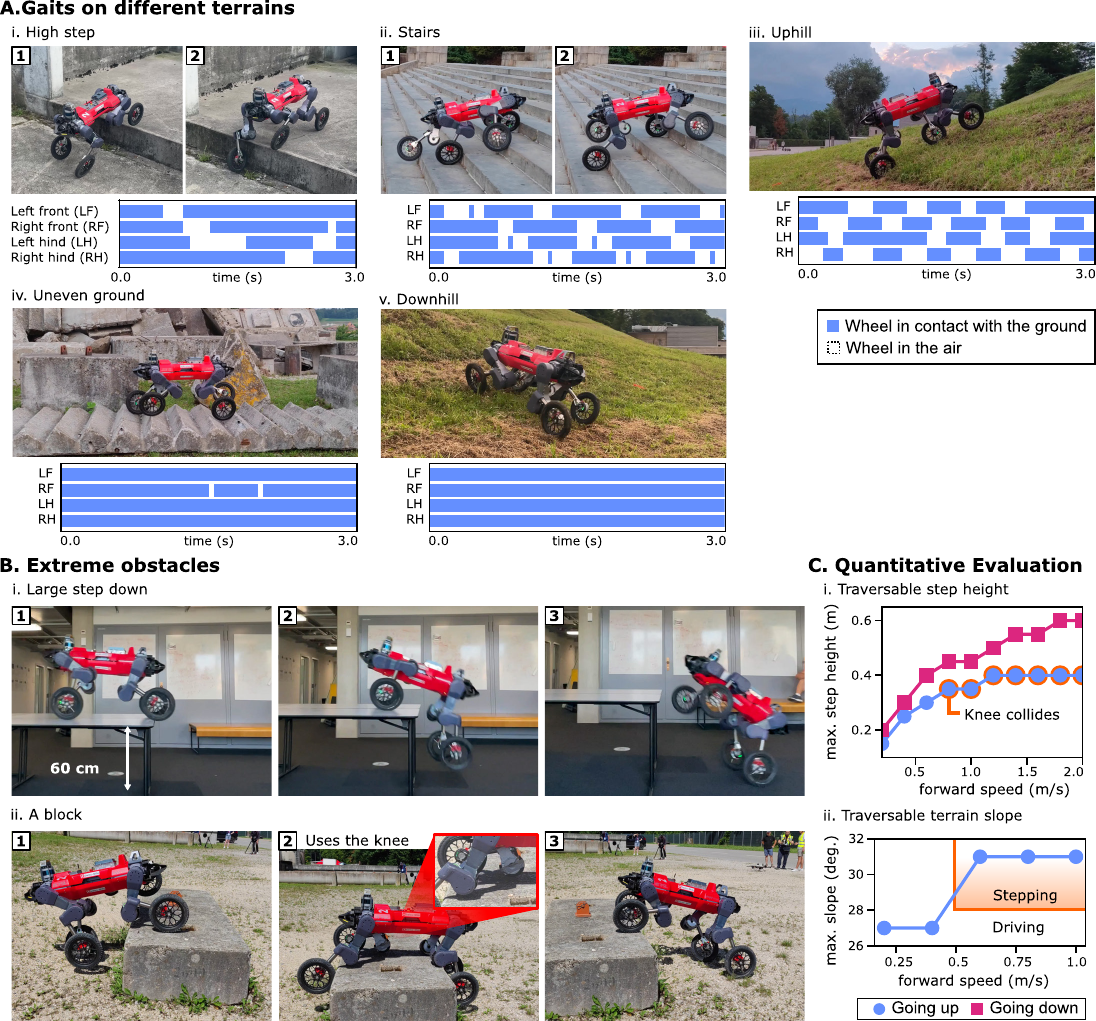}
    \caption{\textbf{Locomotion behaviors in different situations}. \textbf{(A)} Gaits on different terrains. The robot is moving from left to right following target velocity commands given by the joystick (up to \unit[2]{m/s}). The plots show the wheel contact sequences for each terrain. \textbf{(B)} Motion sequences over two extreme obstacles. (i) The robot underwent a full flight phase during the drop while maintaining stability. (ii) When traversing high obstacles, the robot sometimes leveraged other body parts such as its knees.
    \textbf{(C)}  (i) The maximum step height and (ii) the maximum terrain slope traversed by our locomotion controller with a given command velocity, when ascending and descending.
    }
    \label{fig:robot_outdoor}
\end{figure*}

We evaluated our \ac{LLC} over various real-world terrains to observe emerging gaits and assess its robustness.
We provide highlights of our locomotion experiments in \href{https://youtu.be/eEsd10cirqM?list=PLE-BQwvVGf8GOIzjIAeuY6qCVgDGTr5jh}{Movie S3}.
The \ac{LLC} adapts gaits depending on the command velocity and terrain. We tested the policy on various real-world terrains, as illustrated in Fig.~\ref{fig:robot_outdoor}. Our previous \ac{MPC}-based controller~\cite{bjelonic2020whole} lacks robustness and cannot operate in the environments depicted in Fig.~\ref{fig:robot_outdoor}.
Additionally, our controller reached the peak speed of \unit[5.0]{m/s} on flat terrain.
The hardware limit allows for a maximum speed of \unit[6.3]{m/s}, which is determined by the maximum joint speed of \unit[45]{rad/s} multiplied by the wheel radius of \unit[0.14]{m}.

Fig.~\ref{fig:robot_outdoor}A presents distinct behaviors depending on the terrain.
When traversing a large discrete obstacle (Fig.~\ref{fig:robot_outdoor}-i), the robot displayed an asymmetric gait combining creeping~\cite{mcghee1968stability} and driving.
When climbing stairs or steep hills, the robot trotted like a normal point-foot quadruped~\cite{miki2022learning} (Fig.~\ref{fig:robot_outdoor}-ii and Fig.~\ref{fig:robot_outdoor}-iii).
Conversely, the robot drove over the bumpy terrain where the height deviations were comparable to the wheel's radius (Fig.~\ref{fig:robot_outdoor}-iv).
The policy adjusted the reach of each leg to keep the main body stable and kept the wheels in contact with the terrain, acting as an active suspension.
The gait pattern varied depending on the terrain conditions, such as slope or friction. Additionally, the policy adjusted the main body's height based on the situation. For instance, when descending a slope, the policy lowered the body height to enhance stability and prevent tipping over (Fig.~\ref{fig:robot_outdoor}-v).

In Fig.\ref{fig:robot_outdoor}B, we present two scenarios involving high discrete obstacles. 
In Fig.~\ref{fig:robot_outdoor}B-i, we commanded our \ac{LLC} to drive down a table approximately \unit[60]{cm} high. 
As the front legs descended, the robot stretched down its front legs and crouched the hind legs to maintain a level main body.
Once the front legs made contact with the ground, the front wheels rolled forward to regain balance.
In Fig.~\ref{fig:robot_outdoor}B-ii, we show our robot traversing a block of approximately \unit[40]{cm} high. In the middle of the block (ii-2), all the wheels were in the air.
Then the robot crawled forward with its knees until one of the wheels regained contact.
This example shows the advantage of using model-free \ac{RL}~\cite{lee2019robust}.

Quantitative evaluation of the locomotion performance is presented in Fig.\ref{fig:robot_outdoor}C.
In Fig.~\ref{fig:robot_outdoor}C-i, we present the maximum traversable height of the step depending on the command speed. Our robot could traverse higher steps when descending compared to ascending. 
This observation aligns with the results shown in Fig.\ref{fig:obstacles}CD, where our \ac{HLC} avoided high steps to avoid knee collision and ensure safety.
In Fig.~\ref{fig:robot_outdoor}C-ii, we tested \ac{LLC} on slopes with a fixed friction coefficient of 0.7 in simulation. The robot was commanded with a fixed linear velocity to ascend the slope, and success was determined by its ability to climb up for 2 meters.
We observed that the stepping behavior, as depicted in Fig.~\ref{fig:robot_outdoor}A-iii, emerged only on steep slopes with command speeds over \unit[0.5]{m/s}. 
With the stepping gait, the robot was able to climb steeper slopes. This analysis demonstrates the complex characteristics of our \ac{LLC} in terms of gait patterns and traversability. Conventional model-based planning and path-following approaches would struggle to identify and adapt to such complexities.

\subsection*{Comparison to a Conventional Navigation Approach}

\begin{figure*}
    \centering
    \includegraphics{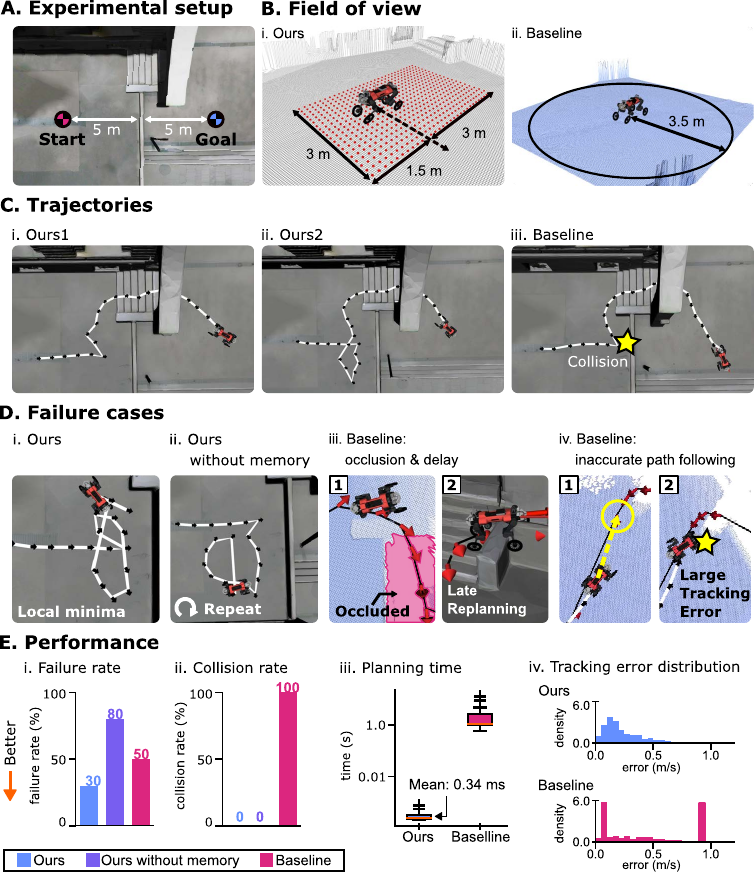}
    \caption{
    \textbf{Comparison to a conventional approach in a point-goal navigation setup.}
    \textbf{(A)} Experimental setup. A goal point is given across the complex obstacle comprising stairs and a wall. The robot is initialized with uniformly sampled yaw angles between -$\pi/4$ to $\pi/4$, facing the goal point.
    \textbf{(B)} Field of views of our approach and the baseline. 
    \textbf{(C)} Trajectories of the two methods. Our method displays two distinct trajectories depending on the initial exploration direction. 
    \textbf{(D)} Failure cases.
    \textbf{(D-i)} Our controller got stuck when the exploration path became longer than its memory capacity.
    \textbf{(D-ii)} Without memory, our approach often fell into local minima.
    \textbf{(D-iii)} Baseline suffered from overconfidence in occlusion and delays in replanning.
    \textbf{(D-iv)} The high pose tracking error of the path-following controller of the baseline led to frequent collisions. 
    \textbf{(E)} Quantitative evaluation of performance. The experiments were repeated 10 times per method.
    \textbf{(E-i, ii)} Failure and collision rates out of the 10 trials.
    \textbf{(E-iii)} Planning time comparison. The error bar denotes one standard deviation. Plus signs indicate outliers.
    \textbf{(E-iv)} Histograms of tracking errors during the experiments. The baseline approach shows two high peaks.
    }
    \label{fig:planner_comp}
\end{figure*}

We compared our approach with the conventional sampling-based navigation planner by Wellhausen et al.~\cite{wellhausen2021rough}. This local navigation planner, used by the Cerberus team in the Subterranean Challenge~\cite{tranzatto2022cerberus}, is designed for normal legged robots.
For both methods, we used the same \ac{LLC}.

We conducted experiments in a point-goal navigation setup, as depicted in Fig.\ref{fig:obstacles}A. The area was scanned with a laser scanner to create a simulation environment, shown in Fig.\ref{fig:planner_comp}A, with fixed start and goal points.

Fig.~\ref{fig:planner_comp}B illustrates the field of view of our \ac{HLC} and the baseline.
To accommodate the limitations during physical deployment, we limited the range of the map to 3.5 meters in both $x$ and $y$ directions. This decision was particularly important when the robot moved at high speeds, reaching up to \unit[2]{m/s}. Using a larger map slowed down the elevation mapping update and resulted in high delay and un-updated regions in the map.

Fig.~\ref{fig:planner_comp}C presents the trajectories of both approaches that successfully reached the goal. Our approach explored the environment until it discovered the staircase. The baseline could also solve the problem when occluded regions were assumed traversable. However, the baseline consistently collided due to the delay issue, which will be explained below.

Fig.\ref{fig:planner_comp}D presents failure cases. Our approach sometimes got stuck after exploring a wide open area (Fig.\ref{fig:planner_comp}D-i). The position memory became full and the agent did not explore further. Additionally, we trained our approach without memory to validate the importance of positional memory. The memoryless policy exhibited repetitive behaviors and struggled to escape local minima (Fig.~\ref{fig:planner_comp}D-ii).

The baseline method faced two challenges: occlusion handling and tracking error of the locomotion controller. Although several heuristics could help mitigate the occlusion problem, the baseline method's ability to handle changing situations was limited due to the delay in re-planning. Concerning the second issue, most existing methods assume perfect tracking; however, the actual locomotion controller experienced delays and tracking errors. Fig.~\ref{fig:planner_comp}D-iv illustrates this issue, where distant pose targets led to high-velocity commands and overshoot.
The robot could not accurately track the next waypoint and collided. This problem becomes more pronounced when dealing with fast-moving robots on rough terrain.

In the quantitative analysis shown in Fig.~\ref{fig:planner_comp}E-i, Our approach with full memory showed the lowest failure rate.
Our method without memory exhibited the highest failure rate. 
The comparison of the failure rate highlights the advantage of exploratory behavior in partially observable scenarios.
Unlike the sampling-based baseline, which is limited to exploring within the provided map, our method enabled the robot to dynamically explore new areas, resulting in a higher success rate. Furthermore, the results obtained from 'Ours without memory' emphasize the importance of the memory mechanism in facilitating effective exploration in static environments.
Notably, only our approach achieved collision-free trajectories (Fig.~\ref{fig:planner_comp}E-ii). This is attributed to the accurate steering capability of our \ac{HLC}, which respects the capabilities of the locomotion policy.

Another benefit of our approach lies in its computational efficiency (Fig.~\ref{fig:planner_comp}E-iii). From updating observations to inferring the neural network, our high-level controller took \unit[0.34]{ms} on average. In contrast, depending on the complexity of the environment, the baseline sometimes required more than a second to update the navigation plan on a desktop machine (AMD Ryzen 9 3950X, GeForce RTX 2080).

The baseline's high failure rate could also be attributed to the imperfect path following. In Fig.~\ref{fig:planner_comp}E-iv, a histogram illustrates the tracking error distributions of both approaches.
The average tracking errors for our approach and the baseline were \unit[0.24]{m/s} and \unit[0.45]{m/s}, respectively.
The baseline exhibited a peak at high tracking error in the histogram, which occurs when there are discrete changes in the command velocity or when the robot is commanded too close to obstacles, causing \ac{LLC} to refuse to follow the command.
In contrast, our high-level controller, trained in conjunction with the low-level controller, demonstrates evenly-distributed tracking error statistics with consistently low tracking error.

\section*{Discussion}

The presented wheeled-legged robot system demonstrates substantial advancements in achieving autonomy and robustness in complex urban environments. The integration of mobility-aware navigation planning and hybrid locomotion contributes to the system's ability to navigate challenging terrain and obstacles while ensuring efficient and fast navigation. 

Our experiments validated the effectiveness of the proposed system in real-world scenarios. Our wheeled-legged robot completed kilometer-scale autonomous missions in urban environments with minimal human intervention.
It navigated through various obstacles such as stairs, irregular steps, natural terrain, and pedestrians.

Our results demonstrate several notable advantages over conventional navigation planning approaches. Firstly, our hierarchical controller actively explores areas beyond its current perception. Unlike traditional sampling-based approaches, our method enables the robot to dynamically explore new areas, improving the success rate. The integration of memory allows the robot to reason about previously visited positions, enhancing its decision-making capabilities in complex environments.

Another major advantage of our approach is its responsiveness. The controller dynamically reacts to unperceived obstacles and effectively navigates through urban environments with pedestrians, continuously adapting to changing situations. The incorporation of real-time data and fast computation enables the robot to leverage up-to-date information, enhancing its ability to navigate challenging terrains and avoid obstacles.

Moreover, the presented hybrid locomotion controller exhibits robustness and versatility in traversing various rough terrain. 
The adaptive gaits observed in our experiments, such as the asymmetric gait for large discrete obstacles, wheel-based locomotion for bumpy terrain, and trotting gait for stairs and steep hills, demonstrate the controller's capability to efficiently traverse diverse terrains.

However, there are still important aspects to consider for future improvements. One such aspect is the incorporation of semantic information into our system. Currently, our system primarily relies on geometric information for navigation, with minimal utilization of semantic information (to adjust the height map for human safety). More advanced scene understanding, such as pavement detection or visual traversability estimation~\cite{wellhausen2020safe}, will allow the robot to make more informed decisions during navigation. This is exemplified by the work of Sorokin et al.~\cite{sorokin2022learning}, where they suggest enhancing a robot's ability to visually differentiate terrains, leading to safer urban navigation.

Another important requirement is fast perception with a wide field of view.
Our \ac{HLC} relies on a limited field of view of up to three meters to the front of the robot. This is inherently limited by using elevation mapping~\cite{miki2022elevation}.
Our system's perception capabilities, although effective for the demonstrated scenarios, may present limitations for faster missions or in environments with high uncertainty.
Our robot hardware is capable of locomotion up to \unit[6.2]{m/s}, but we couldn't demonstrate the maximum speed during autonomous deployment due to the delayed and limited mapping.
Removing terrain elevation mapping and relying on the fast raw sensory stream would be a promising direction for future improvement.

In conclusion, the presented wheeled-legged robot system demonstrates the potential for achieving robust autonomy in complex and dynamic urban environments using data-driven approaches. 
Although challenges remain, such as improving perception capabilities or reducing human labor in map creation, our research paves the way for future advancements in the field of wheeled-legged robots and autonomous urban applications.

Overall, our research contributes to the growing body of knowledge on wheeled-legged robots and autonomous navigation in urban environments. The presented system's robustness, adaptability, and efficiency hold great promise for transforming last-mile delivery and addressing the challenges of urban mobility.

\section*{Materials and Methods}

Our main objective, as depicted in Figure~\ref{fig:intro}B, was to develop a robust control system that enables the robot to navigate along a predefined global path consisting of a sequence of waypoints spaced approximately \unit[2]{m} to \unit[20]{m} apart. The global path can be generated using a graph planner~\cite{kulkarni2022autonomous} or defined manually. It is important to note that although the global planning aspect is essential for the overall navigation process, it is outside the scope of this work.

Due to space constraints, a comprehensive validation of our method is presented in Supplementary Materials.

\subsection*{Overview of the Approach}

Inspired by the existing literature~\cite{nachum2019does, jain2020pixels}, in which hierarchical decomposition of complex tasks enables faster learning and higher performance, we employed \ac{HRL} to extend our previous learning-based velocity tracking controller~\cite{miki2022learning} to waypoint tracking navigation.
In this section, we present an overview of our method, starting with the definition of the hierarchical structure.

\subsubsection*{Defining Hierarchy}\label{hieararchy_def} 

To tackle the waypoint tracking navigation problem, we adopted the two-level \ac{HRL} framework by \cite{nachum2018data}. Various hierarchical structures have been explored in the literature.

Initially, we considered an end-to-end strategy as done by Rudin et al.~\cite{rudin2022advanced}. This method trains a unified policy to simultaneously manage locomotion and navigation tasks, without any hierarchical structure.

An alternative explored in the literature involves a two-level hierarchy, where a high-level policy directs the low-level policy by issuing latent sub-goals at a lower freqency. 
Using learned latent sub-goals for \ac{HRL}~\cite{vezhnevets2017feudal, jain2019hierarchical} offers simplicity and flexibility. There is no need to explicitly define intermediate goals, and the task assignment within the hierarchy is learned.

Our approach instead adopted an explicitly defined sub-goal within a two-level hierarchy. In our setup, the low-level policy focused on locomotion tasks, and the high-level policy focused on navigation by commanding target base velocities to the low-level policy. We opt for explicitly defining sub-goals for practical reasons.

Though the first two approaches could have provided simpler implementations, our decision to explicitly separate the control tasks enabled the independent development of the controllers.
This separation not only simplified collaborative development efforts, allowing teams to work simultaneously on distinct system aspects, but also aligned with common practices in legged robotics. Consequently, this approach facilitated the reuse of pre-trained low-level policies across a range of high-level applications, enhancing the system's versatility and adaptability.

Despite our high-level policy primarily outputing base velocity commands, we also explored commanding gait patterns similarly to Tsounis et al.~\cite{tsounis2020deepgait}. The experiment is described in Supplementary Materials.

\subsubsection*{Training Procedure}
We trained a low-level policy and a high-level policy sequentially.
The low-level policy training involved two stages: teacher policy training followed by student policy training. Then the high-level policy was trained using the trained low-level student policy.

We began by training the teacher policy for the low-level locomotion policy. The teacher policy was trained to follow random velocity targets (and optionally gait parameters) on rough terrains using a \ac{PPO} algorithm~\cite{schulman2017proximal}. In this step, privileged information, including the robot's motion, terrain properties, and noiseless exteroceptive measurements, was utilized to enhance the locomotion performance and convergence of the policy.

Subsequently, the deployable student policy was trained. Unlike the teacher policy, the student policy receives a sequence of noisy and biased \ac{IMU} measurements, joint states, and noisy height scans as input, instead of directly accessing privileged information. Through imitation learning from the teacher policy and leveraging an \ac{RNN} encoder~\cite{miki2022learning}, the student policy was trained to extract features from the temporal data necessary for robust locomotion.

The trained student low-level policy was then regarded as a fixed component, and a high-level navigation policy was trained using a \ac{PPO} algorithm,
The training data was collected in our custom-built simulation environment.
This approach is further explained in the next section.

In addition to the previous three stages, an optional phase of alternating training could be conducted for both policies to enhance their coordination and potentially improve motion smoothness. However, our experiments showed only marginal enhancements from this, and therefore, we did not conduct any further fine-tuning.

\subsection*{Graph-guided Navigation Learning}\label{sec:nav_map}

Navigation graphs, commonly employed in computer games for autonomously navigating characters in synthetic environments~\cite{CryNav, UnrealNav}, played a crucial role in our navigation learning approach.
Inspired by game development, we utilized pre-generated navigation graphs to define initial states, assign feasible paths, and design the reward function during the training of our high-level policy.

\begin{figure*}
    \centering
    \includegraphics[width=\linewidth]{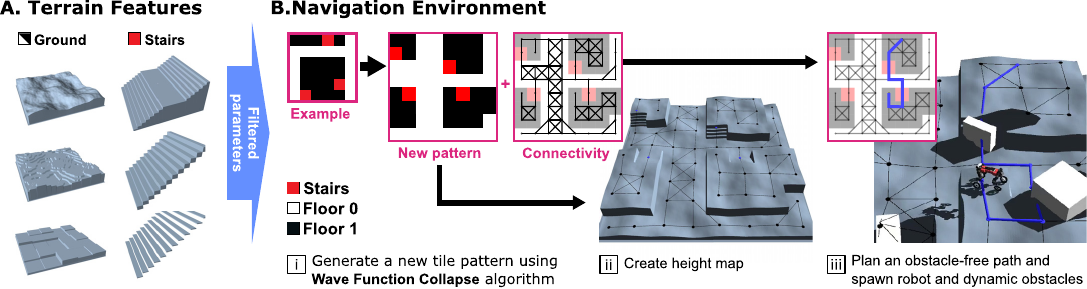}
    \caption{\textbf{Procedural generation of the navigation world.} \textbf{(A)} Filtered parameterized terrain during low-level policy training. \textbf{(B-i)} Generating new tile maps and connectivity graph using the Wave Function Collapse algorithm. \textbf{(B-ii)} Created height map terrain with filtered floor features and stair parameters. \textbf{(B-iii)} Randomly generated navigation path between two nodes provides waypoints during training. Dynamic obstacles (white boxes) are added randomly.}
    \label{fig:terrain_gen}
\end{figure*}

\subsubsection*{World Generation}

Our automatic terrain generation method, illustrated in Fig.~\ref{fig:terrain_gen}, establishes connectivity between different areas of the terrain (tiles), resulting in a navigation graph across the training environment.. For example, tiles with stairs in $x$-direction are exclusively connected to floor tiles along the $x$-axis.

To generate diverse and realistic terrain layouts, we utilized the \ac{WFC} algorithm. This algorithm automatically combines various terrain features such as stairs, floors, and other obstacles. The output of the \ac{WFC} algorithm provided both the composed terrain and the connectivity information between the tiles.

The \ac{WFC} algorithm divides an input tile map (referred to as "Example" in Fig.~\ref{fig:terrain_gen}B) into smaller chunks and rearranges them to create new N by N patterns. This procedural generation approach enabled us to generate a wide variety of navigation worlds with different styles of corridors, rooms, and obstacles.

We defined three types of tiles: Stair, Floor 0, and Floor 1. We provided their relationship to the \ac{WFC} algorithm along with example images. The \ac{WFC} algorithm calculates the probability of each tile type and determines the connectivity to neighboring tile types. By randomly generating tile maps based on these probabilities, we composed the existing tiles, resulting in varied and realistic training environments.
The parameters for the parameterized floor and stairs were selected during the low-level policy training using the terrain filtering algorithm by Lee et al.~\cite{lee2020learning}. See Supplementary Materials for details. 

\subsubsection*{Using Navigation Graphs for RL}

We employed Dijkstra's algorithm~\cite{dijkstra1959note} to find a path between two randomly selected nodes within the graph. Along the graph edge, we sampled two waypoints by interpolating between the robot’s current position projected onto the path and the subsequent graph node, with a fixed look-ahead distance.
The distance was sampled uniformly from [5.0, 20.0]$~\unit{m}$ every episode.
At the end of each path, we included the last node twice as two waypoints. This approach ensured that the agent has clear instructions on the desired trajectory and endpoint.

During the initial training phase, a positive reward was given when the agents moves along the planned path on the graph. 
The reward gradually diminished, and we let the policy train with a sparse reward at the end.
The reward function was defined as follows.

    \begin{equation}
            r_{h, dense} \coloneqq 
        \begin{cases}
        1.0 & |e_{wp^1}|< 0.75 \\
        clip(v\cdot \widehat{e_{wp^1}}, 0.0, v_{thres}) / v_{thres} & \text{otherwise}
        \end{cases}
    \end{equation}
where $e_{wp^1} ={wp}^1 -  p_{robot}$ and $v_{thres} = 0.5$. $p_{robot}$ and ${wp}^1$ denote the positions of the robot and the nearest waypoint, respectively.
    
This reward mechanism encouraged the agent to follow a shortest distance on the navigation graph, minimizing the geodesic distance to the final goal.
The path entailed detours, rather than simply moving straight towards a waypoint.
This approach challenged agents with paths that incorporate tight gaps and sharp turns, thereby pushing their capabilities.

\subsubsection*{Dynamic Obstacles}
In addition to the static structure generated by the \ac{WFC} algorithm, we introduced dynamic obstacles during the training. The dynamic obstacles were randomly placed within the environment and moved towards the robot.

The dynamic obstacles are shown by white boxes in Fig.~\ref{fig:terrain_gen}B-iii. Their number, positions, and velocities were randomly generated each episode.
These obstacles moved towards the robot at speeds ranging from \unit[0.1]{m/s} and \unit[0.5]{m/s}.

\subsection*{High-level Policy Details}\label{sec:learning}

This section is focused on detailing the \ac{MDP} governing the high-level policy ($\pi_{hi}$), encompassing observations and actions. Detailed information about the reward function can be found in supplementary section S3.

\subsubsection*{Observation}

The observation space of $\pi_{hi}$ contains four different modalities.

Firstly, $\pi_{hi}$ observed exterceptive measurements from terrain mapping for obstacle avoidance. The exteroceptive observation followed the definition by Miki et al.\cite{miki2022learning}. We sampled height values around the robot from the robot-centric elevation map~\cite{miki2022elevation}.
Due to limited memory and computational resources onboard, the robot's field of view was restricted to 3 meters to the front and 1.5 meters in other directions. We prioritized shifting the scan pattern towards the front due to the farther perception range afforded by the forward-facing RGB camera.
In addition, the exteroceptive observation included two previous scans taken at \unit[0.1]{s} and \unit[0.2]{s} before to account for the dynamic environments.

Secondly, $\pi_{hi}$ observed the hidden states of the locomotion controller instead of estimated robot states such as gravity vector or twist. Using the hidden state of the \ac{RNN} locomotion policy improved the robustness of our system. This will be explained in detail in Low-level Policy Details. 

To facilitate exploration, we used an additional position buffer. We recorded the visited positions in the world frame at regular intervals of 0.5 meters, along with the corresponding visitation time. The time information included how many time steps the robot stayed in each position. The most recent 20 positions and their respective time information were provided to the policy in the robot frame. 

Lastly, two waypoints were observed. $\pi_{hi}$ observed a short history of two previously given waypoints and three previous outputs of $\pi_{hi}$. This history of waypoints and actions assisted the policy in making a smoother trajectory.

\subsubsection*{Exploration Bonus}

During training, we encouraged the exploration using the explicit position buffer.
This was achieved through an exploration bonus added tp the reward function. 
The exploration bonus, denoted as $r_{exp}$, was calculated as the sum of costs $C(s_t, {wp}^1_t, p^i_{buf})$ over the positions in the buffer $P_{buf}$.

The cost function $C(p_{robot}, {wp}^1, p^i_{buf})$ was defined as 
    \begin{equation}
    C(p_{robot}, {wp}^1, p^i_{buf}) \coloneqq
    \begin{cases}
     0.0  & |p_{robot} - {wp}^1|< 0.75 \\
     - n^i_{buf} & |p_{robot} - p^i_{buf}| < 1.0 \\
    \end{cases}
    \quad .
    \end{equation}
Here, $p_{robot}$ represents the position of the robot, ${wp}^1$ denotes the first waypoint, $p^i_{buf}$ represents the $i$-th position saved in the position buffer, and $n^i_{buf}$ corresponds to the number of visits for the $i$-th position in the position buffer.

In essence, if the robot is not close to the first waypoint and is near a position saved in the position buffer, the agent incurs a penalty proportional to the number of time steps it stayed in that position. This penalty encouraged the agent to explore new areas and prioritize progress towards the first waypoint.

\subsubsection*{Bounded Action Space}

Instead of the commonly used Gaussian action distribution, we used Beta Distribution to represent a bounded action space for the $\pi_{hi}$, as introduced by Chou et al.~\cite{chou2017improving}. This offered several benefits. Firstly, it allowed us to define hard limits on the outputs, enhancing safety and interpretability. Additionally, working with a bounded action space made it easier to regularize the motion and control the behavior of the agent.

Specifically, we defined the bounds of \ac{HLC}'s commands as follows: $v_x \in [-1.0, 2.0]$~\unit{m/s}, $v_y \in [-0.75, 0.75]$~\unit{m/s}, and $\omega_z \in  [-1.25, 1.25]$~\unit{rad/s}. The shift in the $v_x$ range encouraged the policy to consistently face forward during locomotion, aligning with the orientation of the RGB camera mounted on the robot. We provided additional details in Supplementary Materials.

\subsubsection*{Network Architecture}

We employed a combination of architectures tailored for specific input types. For position history, we utilized one dimensional \ac{CNN} layers followed by max pooling, similar to PointNet~\cite{qi2017pointnet}, enabling permutation-invariant processing of spatial information. The height scan around the robot was processed using a 3-layer two dimensional \ac{CNN} layers followed by an \ac{MLP} layer. Other inputs and the output were processed by plain \ac{MLP} layers commonly used for non-spatial data. For the beta distribution parameters, we used the Sigmoid function at the output layer.

\subsection*{Low-level Policy Details}\label{sec:learning2}

The \ac{MDP} for the low-level teacher policy inherited from Miki et al.~\cite{miki2022learning}, with modification to observation and action spaces.
The reward function and the details on the privileged training are provided in Supplementary Materials.

The low-level policy was trained to achieve velocity tracking on random rough terrains. These terrains, designed by Miki et al.\cite{miki2022learning}, are illustrated in Fig.\ref{fig:terrain_gen}A. 
Each terrain type was generated by two to three parameters. During training, we applied the parameter filtering algorithm by Lee et al.~\cite{lee2020learning}. 

The low-level policy is commanded by linear velocity in the x and y direction, as well as yaw rate. 
Linear x velocity is uniformly sampled from [-2.5, 2.5]~\unit[]{m/s}, y velocity from [-1.2, 1.2]~\unit[]{m/s}, and yaw rate from [-1.5, 1.5]~\unit[]{rad/s}. In each episode, a new command is sampled, with a 0.005 probability of random resampling.

\subsubsection*{Observation}

The observation includes three types of information: A sequence of both exteroceptive and proprioceptive measurements, alongside the velocity command.

For the exteroceptive perception, we sampled height values around the robot's wheels from a circular pattern, the same as Miki et al.~\cite{miki2022learning}. 

The proprioception included measurements obtained from body \ac{IMU} and joint encoders.
These measurements convey information about the robot's body acceleration, angular velocities, joint angles, and joint velocities.

As previously discussed, instead of relying on estimated pose and twist by a model-based state estimator as done in several existing works~\cite{hwangbo2019learning, miki2022learning, rudin2022learning}, we directly used \ac{IMU} measurements consisting of linear acceleration and angular velocity. 
This shift was motivated by the observation that conventional state estimators often result in high errors in case of wheel slippage or discrete height changes.
In \href{https://youtu.be/qmgZyh2bfeo?list=PLE-BQwvVGf8GOIzjIAeuY6qCVgDGTr5jh}{Movie S4}, we show a failure case of a locomotion controller due to the state estimation error.

The command was provided as a 3-dimensional vector, including the target base horizontal velocity and target base yaw rate.

\subsubsection*{Privileged Observation}
Privileged observation was only used for teacher policy training. It included noiseless joint states, foot contact state, terrain normal at each foot, foot contact force, robot velocity, and gravity vector in the robot's base frame~\cite{miki2022learning}.

\subsubsection*{Action}
The low-level policy's action is a 16-dimensional vector consisting of joint position commands (12 joints) and wheel velocity commands (4). The joint position and velocity commands were given to the PD controller of each actuator. For a more detailed explanation of the simulation of the actuators, we refer the readers to Supplementary Materials.

In contrast to our prior work~\cite{miki2022learning}, we discarded the use of the \ac{CPG} in the action space to remove any engineered bias in the motion. A detailed comparative study of various action spaces is provided in Supplementary Materials.

\subsubsection*{Network Architecture}
The low-level teacher policy was implemented as a plain three-layer \ac{MLP}, and the low-level student policy was based on the \ac{GRU} architecture by Miki et al.~\cite{miki2022learning}. 

\subsection*{Statistical Analysis}
Statistical analyses were performed using Python. For all the results, we computed mean and standard deviation over the full trajectory using Numpy library. The box plot in Fig.~\ref{fig:planner_comp} is generated using Matplotlib library. 
We collected data at \unit[400]{Hz} using either onboard state estimator or ground truth from simulation.
A low-pass filter with cutoff frequency of \unit[5]{Hz} is applied to the state measurements to reduce high-frequency noise.
The heat map in Fig.~\ref{fig:glatt} is generated by counting the visitation every \unit[1]{m} based on the point cloud localization.
For the \ac{COT} comparison in Kilometer-Scale Autonomous Deployments, we only considered the data points where the linear speed is higher than \unit[0.2]{m/s}.
For the tracking error histogram in Fig.~\ref{fig:planner_comp}, we used data points with command speed higher than \unit[0.5]{m/s}.


\begin{thebibliography}{10} 

\bibitem{hutter2016anymal}
M.~Hutter, C.~Gehring, D.~Jud, A.~Lauber, C.~D. Bellicoso, V.~Tsounis, J.~Hwangbo, K.~Bodie, P.~Fankhauser, M.~Bloesch, others, Anymal-a highly mobile and dynamic quadrupedal robot, {\it 2016 IEEE/RSJ international conference on intelligent robots and systems (IROS)\/},  38--44 (IEEE, 2016).

\bibitem{tranzatto2022cerberus}
M.~Tranzatto, T.~Miki, M.~Dharmadhikari, L.~Bernreiter, M.~Kulkarni, F.~Mascarich, O.~Andersson, S.~Khattak, M.~Hutter, R.~Siegwart, others, Cerberus in the darpa subterranean challenge, {\it Science Robotics\/} p. eabp9742 (2022).

\bibitem{bjelonic2023learning}
F.~Bjelonic, J.~Lee, P.~Arm, D.~Sako, D.~Tateo, J.~Peters, M.~Hutter, Learning-based design and control for quadrupedal robots with parallel-elastic actuators, {\it IEEE Robotics and Automation Letters\/}  1611--1618 (2023).

\bibitem{wellhausen2021rough}
L.~Wellhausen, M.~Hutter, Rough terrain navigation for legged robots using reachability planning and template learning, {\it 2021 IEEE/RSJ International Conference on Intelligent Robots and Systems (IROS)\/},  6914--6921 (IEEE, 2021).

\bibitem{centauro}
N.~Kashiri, L.~Baccelliere, L.~Muratore, A.~Laurenzi, Z.~Ren, E.~M. Hoffman, M.~Kamedula, G.~F. Rigano, J.~Malzahn, S.~Cordasco, P.~Guria, A.~Margan, N.~G. Tsagarakis, Centauro: A hybrid locomotion and high power resilient manipulation platform, {\it IEEE Robotics and Automation Letters\/}  1595--1602 (2019).

\bibitem{bjelonic2019keep}
M.~Bjelonic, C.~D. Bellicoso, Y.~de~Viragh, D.~Sako, F.~D. Tresoldi, F.~Jenelten, M.~Hutter, Keep rollin’—whole-body motion control and planning for wheeled quadrupedal robots, {\it IEEE Robotics and Automation Letters\/}  2116--2123 (2019).

\bibitem{klemm2019ascento}
V.~Klemm, A.~Morra, C.~Salzmann, F.~Tschopp, K.~Bodie, L.~Gulich, N.~K{\"u}ng, D.~Mannhart, C.~Pfister, M.~Vierneisel, others, Ascento: A two-wheeled jumping robot, {\it {ICRA} Int. Conf. on Robotics and Automation (ICRA)\/},  7515--7521 (2019).

\bibitem{Reid2020Mobility}
W.~{Reid}, B.~{Emanuel}, B.~{Chamberlain-Simon}, S.~{Karumanchi}, G.~{Meirion-Griffith}, Mobility mode evaluation of a wheel-on-limb rover on glacial ice analogous to europa terrain, {\it {IEEE} Aerospace Conference\/},  1--9 (2020).

\bibitem{bjelonic2022planning}
M.~Bjelonic, R.~Grandia, M.~Geilinger, O.~Harley, V.~S. Medeiros, V.~Pajovic, E.~Jelavic, S.~Coros, M.~Hutter, Offline motion libraries and online mpc for advanced mobility skills, {\it The International Journal of Robotics Research\/}  903--924 (2022).

\bibitem{bjelonic2020whole}
M.~Bjelonic, R.~Grandia, O.~Harley, C.~Galliard, S.~Zimmermann, M.~Hutter, Whole-body mpc and online gait sequence generation for wheeled-legged robots, {\it 2021 IEEE/RSJ International Conference on Intelligent Robots and Systems (IROS)\/},  8388--8395 (IEEE, 2021).

\bibitem{geilinger2018skaterbots}
M.~Geilinger, R.~Poranne, R.~Desai, B.~Thomaszewski, S.~Coros, Skaterbots: Optimization-based design and motion synthesis for robotic creatures with legs and wheels, {\it ACM Transactions on Graphics (TOG)\/} p. 160 (2018).

\bibitem{hybrid_minicheetah}
M.~Hosseini, D.~Rodriguez, S.~Behnke, State estimation for hybrid locomotion of driving-stepping quadrupeds, {\it 2022 Sixth IEEE International Conference on Robotic Computing (IRC)\/},  103--110 (2022).

\bibitem{bellicoso2018dynamic}
C.~D. Bellicoso, F.~Jenelten, C.~Gehring, M.~Hutter, Dynamic locomotion through online nonlinear motion optimization for quadrupedal robots, {\it IEEE Robotics and Automation Letters\/}  2261--2268 (2018).

\bibitem{jenelten2019dynamic}
F.~Jenelten, J.~Hwangbo, F.~Tresoldi, C.~D. Bellicoso, M.~Hutter, Dynamic locomotion on slippery ground, {\it IEEE Robotics and Automation Letters\/}  4170--4176 (2019).

\bibitem{lee2020learning}
J.~Lee, J.~Hwangbo, L.~Wellhausen, V.~Koltun, M.~Hutter, Learning quadrupedal locomotion over challenging terrain, {\it Science robotics\/} {\bf 5} (2020).

\bibitem{miki2022learning}
T.~Miki, J.~Lee, J.~Hwangbo, L.~Wellhausen, V.~Koltun, M.~Hutter, Learning robust perceptive locomotion for quadrupedal robots in the wild, {\it Science Robotics\/} p. eabk2822 (2022).

\bibitem{xi2016selecting}
W.~Xi, Y.~Yesilevskiy, C.~D. Remy, Selecting gaits for economical locomotion of legged robots, {\it The International Journal of Robotics Research\/}  1140--1154 (2016).

\bibitem{yang2022fast}
Y.~Yang, T.~Zhang, E.~Coumans, J.~Tan, B.~Boots, Fast and efficient locomotion via learned gait transitions, {\it Conference on Robot Learning\/},  773--783 (PMLR, 2022).

\bibitem{bellegarda2019trajectory}
G.~Bellegarda, K.~Byl, Trajectory optimization for a wheel-legged system for dynamic maneuvers that allow for wheel slip, {\it 2019 IEEE 58th Conference on Decision and Control (CDC)\/},  7776--7781 (IEEE, 2019).

\bibitem{wellhausen2022ART}
L.~Wellhausen, M.~Hutter, Artplanner: Robust legged robot navigation in the field, {\it Field Robotics\/},  413--434 (2023).

\bibitem{frey2022locomotion}
J.~Frey, D.~Hoeller, S.~Khattak, M.~Hutter, Locomotion policy guided traversability learning using volumetric representations of complex environments, {\it 2022 IEEE/RSJ International Conference on Intelligent Robots and Systems (IROS)\/},  5722--5729 (IEEE, 2022).

\bibitem{chavez2018learning}
R.~O. Chavez-Garcia, J.~Guzzi, L.~M. Gambardella, A.~Giusti, Learning ground traversability from simulations, {\it IEEE Robotics and Automation letters\/}  1695--1702 (2018).

\bibitem{vapnik2015learning}
V.~Vapnik, R.~Izmailov, Learning using privileged information: Similarity control and knowledge transfer, {\it Journal of Machine Learning Research\/}  2023--2049 (2015).

\bibitem{chen2020learning}
D.~Chen, B.~Zhou, V.~Koltun, P.~Kr{\"a}henb{\"u}hl, Learning by cheating, {\it Conference on Robot Learning\/},  66--75 (PMLR, 2020).

\bibitem{snider2009pure}
J.~M. Snider, others, Automatic steering methods for autonomous automobile path tracking, {\it Robotics Institute, Pittsburgh, PA, Tech. Rep. CMU-RITR-09-08\/}  (2009).

\bibitem{miki2022elevation}
T.~Miki, L.~Wellhausen, R.~Grandia, F.~Jenelten, T.~Homberger, M.~Hutter, Elevation mapping for locomotion and navigation using gpu, {\it 2022 IEEE/RSJ International Conference on Intelligent Robots and Systems (IROS)\/},  2273--2280 (IEEE, 2022).

\bibitem{ji2022concurrent}
G.~Ji, J.~Mun, H.~Kim, J.~Hwangbo, Concurrent training of a control policy and a state estimator for dynamic and robust legged locomotion, {\it IEEE Robotics and Automation Letters\/}  (2022).

\bibitem{CryNav}
{CryEngine}, Ai and navigation system - cryengine 5 documentation, \url{https://docs.cryengine.com/pages/viewpage.action?pageId=26869983}. [Online; accessed August-2022].

\bibitem{UnrealNav}
{Unreal Engine}, Navigation system - unreal engine 5 documentation, \url{https://docs.unrealengine.com/5.0/en-US/navigation-system-in-unreal-engine/}. [Online; accessed August-2022].

\bibitem{WFC}
M.~Gumin, Wave function collapse algorithm, \url{https://github.com/mxgmn/} (2016).

\bibitem{dijkstra1959note}
E.~DIJKSTRA, A note on two problems in connexion with graphs, {\it Numerische Mathematik\/}  269--271 (1959).

\bibitem{jelavic2022open3d}
E.~Jelavic, J.~Nubert, M.~Hutter, Open3d slam: Point cloud based mapping and localization for education, {\it Robotic Perception and Mapping: Emerging Techniques, ICRA 2022 Workshop\/}, p.~24 (ETH Zurich, Robotic Systems Lab, 2022).

\bibitem{anymal_in_darpa}
{Robotic Systems Lab}, Team cerberus wins the darpa subterranean challenge, \url{https://youtu.be/fCHOU-fw2c0?si=LjksAckgpSwfMqTC} (2023). [Online; accessed March-2024].

\bibitem{hwangbo2019learning}
J.~Hwangbo, J.~Lee, A.~Dosovitskiy, D.~Bellicoso, V.~Tsounis, V.~Koltun, M.~Hutter, Learning agile and dynamic motor skills for legged robots, {\it Science Robotics\/} p. eaau5872 (2019).

\bibitem{seok2014design}
S.~Seok, A.~Wang, M.~Y. Chuah, D.~J. Hyun, J.~Lee, D.~M. Otten, J.~H. Lang, S.~Kim, Design principles for energy-efficient legged locomotion and implementation on the mit cheetah robot, {\it Ieee/asme transactions on mechatronics\/}  1117--1129 (2014).

\bibitem{mcghee1968stability}
R.~B. McGhee, A.~A. Frank, On the stability properties of quadruped creeping gaits, {\it Mathematical Biosciences\/}  331--351 (1968).

\bibitem{lee2019robust}
J.~Lee, J.~Hwangbo, M.~Hutter, Robust recovery controller for a quadrupedal robot using deep reinforcement learning, {\it arXiv preprint arXiv:1901.07517\/}  (2019).

\bibitem{wellhausen2020safe}
L.~Wellhausen, R.~Ranftl, M.~Hutter, Safe robot navigation via multi-modal anomaly detection, {\it IEEE Robotics and Automation Letters\/}  1326--1333 (2020).

\bibitem{sorokin2022learning}
M.~Sorokin, J.~Tan, C.~K. Liu, S.~Ha, Learning to navigate sidewalks in outdoor environments, {\it IEEE Robotics and Automation Letters\/}  3906--3913 (2022).

\bibitem{kulkarni2022autonomous}
M.~Kulkarni, M.~Dharmadhikari, M.~Tranzatto, S.~Zimmermann, V.~Reijgwart, P.~De~Petris, H.~Nguyen, N.~Khedekar, C.~Papachristos, L.~Ott, others, Autonomous teamed exploration of subterranean environments using legged and aerial robots, {\it 2022 International Conference on Robotics and Automation (ICRA)\/},  3306--3313 (IEEE, 2022).

\bibitem{nachum2019does}
O.~Nachum, H.~Tang, X.~Lu, S.~Gu, H.~Lee, S.~Levine, Why does hierarchy (sometimes) work so well in reinforcement learning?, {\it arXiv preprint arXiv:1909.10618\/}  (2019).

\bibitem{jain2020pixels}
D.~Jain, K.~Caluwaerts, A.~Iscen, From pixels to legs: Hierarchical learning of quadruped locomotion, {\it Proceedings of the 2020 Conference on Robot Learning\/}, J.~Kober, F.~Ramos, C.~Tomlin, eds.,  91--102 (PMLR, 2021).

\bibitem{nachum2018data}
O.~Nachum, S.~S. Gu, H.~Lee, S.~Levine, Data-efficient hierarchical reinforcement learning, {\it Advances in neural information processing systems\/} {\bf 31} (2018).

\bibitem{rudin2022advanced}
N.~Rudin, D.~Hoeller, M.~Bjelonic, M.~Hutter, Advanced skills by learning locomotion and local navigation end-to-end, {\it 2022 IEEE/RSJ International Conference on Intelligent Robots and Systems (IROS)\/},  2497--2503 (IEEE, 2022).

\bibitem{vezhnevets2017feudal}
A.~S. Vezhnevets, S.~Osindero, T.~Schaul, N.~Heess, M.~Jaderberg, D.~Silver, K.~Kavukcuoglu, Feudal networks for hierarchical reinforcement learning, {\it International Conference on Machine Learning\/},  3540--3549 (PMLR, 2017).

\bibitem{jain2019hierarchical}
D.~Jain, A.~Iscen, K.~Caluwaerts, Hierarchical reinforcement learning for quadruped locomotion, {\it 2019 IEEE/RSJ International Conference on Intelligent Robots and Systems (IROS)\/},  7551--7557 (IEEE, 2019).

\bibitem{tsounis2020deepgait}
V.~Tsounis, M.~Alge, J.~Lee, F.~Farshidian, M.~Hutter, Deepgait: Planning and control of quadrupedal gaits using deep reinforcement learning, {\it IEEE Robotics and Automation Letters\/}  3699--3706 (2020).

\bibitem{schulman2017proximal}
J.~Schulman, F.~Wolski, P.~Dhariwal, A.~Radford, O.~Klimov, Proximal policy optimization algorithms, {\it arXiv preprint arXiv:1707.06347\/}  (2017).

\bibitem{chou2017improving}
P.-W. Chou, D.~Maturana, S.~Scherer, Improving stochastic policy gradients in continuous control with deep reinforcement learning using the beta distribution, {\it International conference on machine learning\/},  834--843 (PMLR, 2017).

\bibitem{qi2017pointnet}
C.~R. Qi, H.~Su, K.~Mo, L.~J. Guibas, Pointnet: Deep learning on point sets for 3d classification and segmentation, {\it Proceedings of the IEEE conference on computer vision and pattern recognition\/},  652--660 (2017).

\bibitem{rudin2022learning}
N.~Rudin, D.~Hoeller, P.~Reist, M.~Hutter, Learning to walk in minutes using massively parallel deep reinforcement learning, {\it Conference on Robot Learning\/},  91--100 (PMLR, 2022).

\bibitem{hagberg2008networkx}
A.~A. Hagberg, D.~A. Schult, P.~J. Swart, Exploring network structure, dynamics, and function using networkx, {\it Proceedings of the 7th Python in Science Conference\/}, G.~Varoquaux, T.~Vaught, J.~Millman, eds.,  11 -- 15 (Pasadena, CA USA, 2008).

\bibitem{muller2018driving}
M.~Mueller, A.~Dosovitskiy, B.~Ghanem, V.~Koltun, Driving policy transfer via modularity and abstraction, {\it Proceedings of The 2nd Conference on Robot Learning\/}, A.~Billard, A.~Dragan, J.~Peters, J.~Morimoto, eds.,  1--15 (PMLR, 2018).

\bibitem{agarwal2023legged}
A.~Agarwal, A.~Kumar, J.~Malik, D.~Pathak, Legged locomotion in challenging terrains using egocentric vision, {\it Conference on Robot Learning\/},  403--415 (PMLR, 2023).

\bibitem{kahn2021badgr}
G.~Kahn, P.~Abbeel, S.~Levine, Badgr: An autonomous self-supervised learning-based navigation system, {\it IEEE Robotics and Automation Letters\/}  1312--1319 (2021).

\bibitem{kahn2021land}
G.~Kahn, P.~Abbeel, S.~Levine, Land: Learning to navigate from disengagements, {\it IEEE Robotics and Automation Letters\/}  1872--1879 (2021).

\bibitem{kim2022learning}
Y.~Kim, C.~Kim, J.~Hwangbo, Learning forward dynamics model and informed trajectory sampler for safe quadruped navigation, {\it Conference on Robotics-Science and Systems (RSS)\/} (Robotics: Science and Systems Foundation, 2022).

\bibitem{truong2021learning}
J.~Truong, D.~Yarats, T.~Li, F.~Meier, S.~Chernova, D.~Batra, A.~Rai, Learning navigation skills for legged robots with learned robot embeddings, {\it 2021 IEEE/RSJ International Conference on Intelligent Robots and Systems (IROS)\/},  484--491 (IEEE, 2021).

\bibitem{hoeller2021learning}
D.~Hoeller, L.~Wellhausen, F.~Farshidian, M.~Hutter, Learning a state representation and navigation in cluttered and dynamic environments, {\it IEEE Robotics and Automation Letters\/}  5081--5088 (2021).

\bibitem{pfeiffer2018reinforced}
M.~Pfeiffer, S.~Shukla, M.~Turchetta, C.~Cadena, A.~Krause, R.~Siegwart, J.~Nieto, Reinforced imitation: Sample efficient deep reinforcement learning for mapless navigation by leveraging prior demonstrations, {\it IEEE Robotics and Automation Letters\/}  4423--4430 (2018).

\bibitem{manderson2020vision}
T.~Manderson, J.~C.~G. Higuera, S.~Wapnick, J.~Tremblay, F.~Shkurti, D.~Meger, G.~Dudek, Vision-based goal-conditioned policies for underwater navigation in the presence of obstacles, {\it Conference on Robotics-Science and Systems (RSS)\/} (Robotics: Science and Systems Foundation, 2020).

\bibitem{wang2023diffusion}
H.~Wang, S.~Chen, S.~Sun, Diffusion model-augmented behavioral cloning, {\it CoRR\/} {\bf abs/2302.13335} (2023).

\bibitem{savinov2018semi}
N.~Savinov, A.~Dosovitskiy, V.~Koltun, Semi-parametric topological memory for navigation, {\it International Conference on Learning Representations\/} (2018).

\bibitem{fu2022coupling}
Z.~Fu, A.~Kumar, A.~Agarwal, H.~Qi, J.~Malik, D.~Pathak, Coupling vision and proprioception for navigation of legged robots, {\it Proceedings of the IEEE/CVF Conference on Computer Vision and Pattern Recognition\/},  17273--17283 (2022).

\bibitem{wijmans2019dd}
E.~Wijmans, A.~Kadian, A.~S. Morcos, S.~Lee, I.~Essa, D.~Parikh, M.~Savva, D.~Batra, Dd-ppo: Learning near-perfect pointgoal navigators from 2.5 billion frames, {\it International Conference on Learning Representations\/} (2019).

\bibitem{choi2019deep}
J.~Choi, K.~Park, M.~Kim, S.~Seok, Deep reinforcement learning of navigation in a complex and crowded environment with a limited field of view, {\it 2019 International Conference on Robotics and Automation (ICRA)\/},  5993--6000 (IEEE, 2019).

\bibitem{wijmans2023emergence}
E.~Wijmans, M.~Savva, I.~Essa, S.~Lee, A.~S. Morcos, D.~Batra, Emergence of maps in the memories of blind navigation agents, {\it International Conference on Learning Representations\/} (2023).

\bibitem{zhu2021deep}
K.~Zhu, T.~Zhang, Deep reinforcement learning based mobile robot navigation: A review, {\it Tsinghua Science and Technology\/}  674--691 (2021).

\bibitem{surmann2020deep}
H.~Surmann, C.~Jestel, R.~Marchel, F.~Musberg, H.~Elhadj, M.~Ardani, Deep reinforcement learning for real autonomous mobile robot navigation in indoor environments, {\it arXiv preprint arXiv:2005.13857\/}  (2020).

\bibitem{yang2021learning}
R.~Yang, M.~Zhang, N.~Hansen, H.~Xu, X.~Wang, Learning vision-guided quadrupedal locomotion end-to-end with cross-modal transformers, {\it Deep RL Workshop NeurIPS 2021\/} (2021).

\bibitem{anderson2018evaluation}
P.~Anderson, A.~Chang, D.~S. Chaplot, A.~Dosovitskiy, S.~Gupta, V.~Koltun, J.~Kosecka, J.~Malik, R.~Mottaghi, M.~Savva, others, On evaluation of embodied navigation agents, {\it CoRR\/} {\bf abs/1807.06757} (2018).

\bibitem{gangapurwala2022rloc}
S.~Gangapurwala, M.~Geisert, R.~Orsolino, M.~Fallon, I.~Havoutis, Rloc: Terrain-aware legged locomotion using reinforcement learning and optimal control, {\it IEEE Transactions on Robotics\/}  (2022).

\bibitem{kolvenbach2018efficient}
H.~Kolvenbach, D.~Bellicoso, F.~Jenelten, L.~Wellhausen, M.~Hutter, Efficient gait selection for quadrupedal robots on the moon and mars, {\it 14th International Symposium on Artificial Intelligence, Robotics and Automation in Space (i-SAIRAS 2018)\/} (ESA Conference Bureau, 2018).

\bibitem{ward2019improving}
P.~N. Ward, A.~Smofsky, A.~J. Bose, Improving exploration in soft-actor-critic with normalizing flows policies, {\it arXiv preprint arXiv:1906.02771\/}  (2019).

\bibitem{zhou2020plas}
W.~Zhou, S.~Bajracharya, D.~Held, Plas: Latent action space for offline reinforcement learning, {\it Conference on Robot Learning\/},  1719--1735 (PMLR, 2021).

\bibitem{allshire2021laser}
A.~Allshire, R.~Mart{\'\i}n-Mart{\'\i}n, C.~Lin, S.~Manuel, S.~Savarese, A.~Garg, Laser: Learning a latent action space for efficient reinforcement learning, {\it 2021 IEEE International Conference on Robotics and Automation (ICRA)\/},  6650--6656 (IEEE, 2021).

\bibitem{kingma2013auto}
D.~P. Kingma, M.~Welling, Auto-encoding variational bayes, {\it 2nd International Conference on Learning Representations\/}  (2014).

\bibitem{dinh2016density}
L.~Dinh, J.~Sohl-Dickstein, S.~Bengio, Density estimation using real nvp, {\it International Conference on Learning Representations\/}  (2016).

\bibitem{brant2017minimal}
J.~C. Brant, K.~O. Stanley, Minimal criterion coevolution: a new approach to open-ended search, {\it Proceedings of the Genetic and Evolutionary Computation Conference\/},  67--74 (2017).

\bibitem{team2021open}
O.~E.~L. Team, A.~Stooke, A.~Mahajan, C.~Barros, C.~Deck, J.~Bauer, J.~Sygnowski, M.~Trebacz, M.~Jaderberg, M.~Mathieu, N.~McAleese, N.~Bradley{-}Schmieg, N.~Wong, N.~Porcel, R.~Raileanu, S.~Hughes{-}Fitt, V.~Dalibard, W.~M. Czarnecki, Open-ended learning leads to generally capable agents, {\it CoRR\/} {\bf abs/2107.12808} (2021).

\bibitem{loquercio2021learning}
A.~Loquercio, E.~Kaufmann, R.~Ranftl, M.~M{\"u}ller, V.~Koltun, D.~Scaramuzza, Learning high-speed flight in the wild, {\it Science Robotics\/} p. eabg5810 (2021).

\bibitem{ross2011reduction}
S.~Ross, G.~Gordon, D.~Bagnell, A reduction of imitation learning and structured prediction to no-regret online learning, {\it Proceedings of the fourteenth international conference on artificial intelligence and statistics\/},  627--635 (JMLR Workshop and Conference Proceedings, 2011).

\end{thebibliography}

\section*{Acknowledgments}
\textbf{Funding}:
This work was supported by the Mobility Initiative grant funded through the ETH Zurich Foundation, European Union’s Horizon 2020 research and innovation programme under grant agreement number 101070405 and 101016970, Swiss National Science Foundation through the National Centre of Competence in Digital Fabrication (NCCR dfab), European Union's Horizon Europe Framework Programme under grant agreement number 852044 and 101070596, and Apple Inc.
\textbf{Author contribution}:
J.L. conceived the main idea for the approach and was responsible for the implementation and training of the controllers. The high-level policy was trained by J.L. while M.B. developed the simulation environment for the low-level policy and also trained it. The navigation system and safety layer were collaboratively devised and implemented by J.L., M.B., A.R., and L.W. Real-world experiments were planned and executed with contributions from A.R. and L.W. Initial setup of the low-level controller was facilitated by T.M. All authors contributed to system integration and experimentation.
\textbf{Other contributors}:
We appreciate Turcan Tuna for helping us integrate Open3D SLAM, Julian Keller for the API integration, and Giorgio Valsecchi for the hardware support.
\textbf{Conflict of interest}: The authors declare that they have no competing interests.
\textbf{Data and materials availability}: All data needed to evaluate the conclusions in the paper are present in the main text or the Supplementary Materials.
The datasets and codes to generate Fig. 3, Fig. 6, and Fig. 7 are available at DOI: 10.5061/dryad.gxd2547tg

\clearpage

\noindent

\clearpage
\setcounter{table}{0}
\makeatletter 
\renewcommand{\thetable}{S\@arabic\c@table}
\makeatother

\setcounter{figure}{0}
\makeatletter 
\renewcommand{\thefigure}{S\@arabic\c@figure}
\makeatother

\newpage


\newpage
\section*{Supplementary Materials}
\subsection*{Nomenclature}
\makebox[1.2cm]{$\hat{(\cdot)}$} normalized vector\\
\makebox[1.2cm]{${(\cdot)}_{des}$} desired quantity\\
\makebox[1.2cm]{$v$} linear velocity of the robot in world frame\\
\makebox[1.2cm]{$v^B$} linear velocity of the robot in $B$ frame\\
\makebox[1.2cm]{${p}_{A}$} position of $A$ in world frame \\
\makebox[1.2cm]{${wp}^1$} first waypoint position \\
\makebox[1.2cm]{${wp}^2$} second waypoint position\\
\makebox[1.2cm]{$P_{buf}$} the position buffer \\
\makebox[1.2cm]{$p^i_{buf}$} the $i$-th position saved in the position buffer \\
\makebox[1.2cm]{$n^i_{buf}$} The number of visits for the $i$-th position saved in the position buffer \\
\makebox[1.2cm]{$\omega$} angular velocity\\
\makebox[1.2cm]{$\tau$} joint torque\\
\makebox[1.3cm]{$q$} joint position\\
\makebox[1.2cm]{$\psi$} yaw angle\\
\makebox[1.2cm]{$r_f$} linear position of a foot\\
\makebox[1.2cm]{$e_g$} gravity vector\\
\makebox[1.2cm]{$I_{c,body}$} index set of body contacts\\
\makebox[1.2cm]{$I_{c,wheel}$} index set of wheel contacts\\
\makebox[1.2cm]{$\lvert \cdot \rvert$} cardinality of a set or $l_1$ norm\\
\makebox[1.2cm]{$\lvert\lvert \cdot \rvert\rvert$} $l_2$ norm\\

\subsection*{Implementation Details}

In this section, we add some implementation details that are necessary to enable kilometer-scale autonomous deployments and a successful sim-to-real transfer.

\subsubsection*{Localization}
Using Open3D SLAM~\cite{jelavic2022open3d}, we localize based on the data of the Velodyne VLP-16 LiDAR mounted on top of the robot in a known point cloud map generated with the Leica BLK2GO reality capture device.
Notably, Open3D SLAM only uses well-known algorithms, such as \ac{ICP}, in their core form. The Open3D library is used as a backend for 3D data processing.
For this work, we adapted Open3D SLAM to use odometry from \ac{IMU} and joint encoder data as prior for scan-to-map matching.
We expect this high-frequency odometry source to contribute to the systems robustness because scan-to-scan matching might fail at higher speeds or in degenerate environments.

\subsubsection*{Global Planning}
For global navigation, we also rely on well-established methods and libraries. 
Leveraging the reality capture data, we manually design a sparse navigation graph offline as shown in Fig.~\ref{fig:glatt}.
During deployment we compute the shortest path from the robot's current position to the goal position with Dijkstra using the networkX~\cite{hagberg2008networkx} Python library.

\subsubsection*{Waypoint Selection}
The global path obtained from the navigation graph can contain nodes that are tens of meters apart although the high-level policy was trained with waypoints that are at most \unit[20]{m} apart.
To resolve this issue, we introduce a simple waypoint selection method called "anchor pursuit", inspired by the "pure pursuit" path following algorithm.
If the next path nodes is less than \unit[3]{m} from the current robot position, we select it as a waypoint.
In case it is farther, we instead project the robot position onto the path and select a waypoint with a \unit[3]{m} lookahead distance.
This ensures that the nodes or so-called anchor points are always approached and the robot doesn't take undesired shortcuts around these core waypoints of the global path.
Also note that this is different from interpolating the global path at fixed distances and sequentially approaching these sub-waypoints since with "anchor pursuit" the sub-waypoints are moving forward with the robot.
This allows for greater freedom in circumnavigating obstacles, since interpolated waypoints do not need to be followed exactly.
In future work, one could add exteroceptive information such as semantically segmented images~\cite{muller2018driving}, to generate more sophisticated refined paths for the hierarchical controller or even incorporate this capability into the high-level policy itself.

\subsubsection*{Local Terrain Mapping}
To obtain the extereoceptive observations on real hardware, we use a GPU-accelerated geometric terrain mapping approach~\cite{miki2022elevation}, which provides a local elevation map around the robot based on the data of the two Robosense RS-Bpearl dome LiDARs mounted at the front and rear of the robot.
The points for the foot scan for the locomotion policy and the base scan for the navigation policy are extracted from the elevation map.
Removing the local terrain mapping and directly providing the raw extereoceptive data, such as depth images~\cite{agarwal2023legged} or point clouds, is subject to future work and can help to reduce processing delays.

\subsubsection*{Human Detection}
For human detection, we use the Stereolabs ZED 2i stereo camera.
Their software development kit provides a "Spatial Object Detection" feature, which detects humans and provides an estimate of the human's position in the camera frame.
As shown in Fig.~\ref{fig:obstacles}B, this position is used to augment the foot and base scans with a safety margin around the human.

\subsubsection*{Modeling Actuators}
For successful sim-to-real transfer, it is crucial to simulate the dynamics of the joint actuators. The non-linear characteristics of robotic actuators such as joint friction, delay, and backlash are very difficult to model with simple analytic models. Instead of modeling the whole actuator, we used neural network models to simulate the complex dynamics efficiently. 

The joint actuators are fully simulated by an actuator network~\cite{hwangbo2019learning}, which is trained with accurate torque measurements from \ac{SEA}s.
Wheel actuators are pseudo-direct drives and we do not have access to accurate torque measurements. 
Instead, we learn the mapping from velocity command and a history of past velocity readings to the motor current, which is
\begin{equation}
I_t = f(\dot{\phi}_{target}| \dot{\phi}_{t-1}, \dot{\phi}_{t-2}, \cdots),    
\end{equation} using a neural network. 
Then, the torque is computed by
\begin{equation}
    \tau_t = K_\tau * GR* I_t 
\end{equation}
 where $K_\tau$ is the torque constant and $GR$ is the gear ratio.
 Additionally, we simulated joint friction such that 
 \begin{equation}
      \tau_t = K_\tau * GR* I_t + \tau_{friction}.
 \end{equation}
 The friction is modeled by two terms. Coulomb friction 
 \begin{equation}
     \tau_{friction, C} = - C_1 \dot{\phi}
 \end{equation}
 and the stick friction 
 \begin{equation}
     \tau_{friction, S} = - C_2 sgn(\dot{\phi})
 \end{equation}
 with randomized constants $C_1$ and $C_2$. $C_1$ and $C_2$. The constants were included in the privileged observation.

\subsection*{Reward Functions}
In this section, we provide a detailed explanation on the reward function for each agent.
We categorize the reward functions into three groups.
high-level policy reward~($r_{h}$), low-level policy reward~($r_{l}$) for following the commands given by the $\pi_{hi}$, and the regularization reward~($r_{r}$) which consists of constraint-related objectives and regularization terms.

The main objective of the $\pi_{hi}$ is defined by $r_{h}$, and the $\pi_{lo}$ focuses on the low-level control, such as pose control, balancing, and locomotion control, by maximizing the discounted sum of $r_{l}$. 
$r_{r}$ defines additional sub-objectives such as joint velocity penalty, torque minimization, or action smoothness.
Such regularization objectives are often introduced for robotic applications to avoid damaging hardware or to facilitate sim-to-real transfer. In \ac{RL}, it is done by adding $r_r$ to the reward functions~\cite{lee2020learning, rudin2022learning, ji2022concurrent}.

The low-level policy is trained using $r_{l} +  r_r$, and the high-level policy is trained using $r_h + w_{l} \cdot ( r_{l} + r_r )$ with a constant scale $w_{l}$. We chose the value of $w_{l}$ such that the expected sum of $r_h$ and $r_l + r_r$ are at a similar magnitude. This makes $r_{h}$ generate smooth and low-effort trajectories that respect the capability of the low-level policy.

\subsubsection*{High-level Policy Rewards}
The high-level reward ($r_h$) is defined as a linear combination of functions below.

A goal reaching reward is defined as a sparse reward for reaching the first waypoint (${wp}^1$):
    \begin{equation}
        r_{h, goal} \coloneqq
        \begin{cases}
        1.0 & |p_{robot} - {wp}^1|< 0.75 \\
        0.0 & \text{otherwise}
        \end{cases}.
    \end{equation}
In addition to the sparse reward, we used a dense reward to accelearate learning at the beginning:
    \begin{equation}
            r_{h, dense} \coloneqq 
        \begin{cases}
        1.0 & |e_{wp^1}|< 0.75 \\
        clip(v\cdot \widehat{e_{wp^1}}, 0.0, v_{thres}) / v_{thres} & \text{otherwise}
        \end{cases}
    \end{equation}
    where $e_{wp^1} = p_{robot} - {wp}^1$ and $v_{thres} = 0.5$.
    
An exploration bonus as explained in the main text:
    \begin{equation}
        r_{h, exp} \coloneqq  \sum_{P_{buf}} C(s_t, {wp}^1_t, p^i_{buf}),
    \end{equation}
    where 
    \begin{equation}
    C(p_{robot}, {wp}^1, p^i_{buf}) \coloneqq
    \begin{cases}
     0.0  & |p_{robot} - {wp}^1|< 0.75 \\
     - n^i_{buf} & |p_{robot} - p^i_{buf}| < 1.0 \\
    \end{cases}.
    \end{equation}

Additionally, we defined a near-goal stability reward. This reward motivates the robot to stay still near the goal point. This reward is a part of regularization reward, but this is only active during the high-level policy training:
    
    \begin{equation}
        r_{h, stability} \coloneqq
        \begin{cases}
        \exp{(-2.0 \lvert\lvert v \rvert\rvert ^2 )} & |p_{robot} - {wp}^1|< 0.75 \\
        0.0 & \text{otherwise}
        \end{cases}.
    \end{equation}

\subsubsection*{Low-level Policy Rewards}
$r_l$ is modified from the reward terms by Miki et al.~\cite{miki2022learning}.
$r_l$ is defined with the linear combination of the following reward terms.

The linear velocity tracking reward encourages the policy to follow a desired horizontal velocity (velocity in $xy$ plane) command:
    \begin{equation}
    r_{lv} \coloneqq 
    \begin{cases}
    2.0\exp(- 2.0 \cdot \lvert\lvert {v_{ xy}^{body}}  \rvert\rvert^2), & \text{if } |{v}_{des}| < 0.05 \\
    \exp(-2.0  \lvert\lvert{v_{xy}^{body}} - {v}_{des}\rvert\rvert^2) + {v}_{des} \cdot {v_{xy}^{body}},& \text{otherwise}
     \end{cases},
\end{equation}
where ${v}_{des} \in \mathbb{R}^2$ is the desired horizontal velocity.

We also defined a reward to encourage the policy to follow a desired yaw velocity command:
    \begin{equation}
    r_{av} \coloneqq \exp(- 2.0(\omega_z^{body} - \omega_{des})^2).
\end{equation}
As we aim for stable base motions, we defined a penalty for the body velocity in directions not part of the command:
    \begin{equation}
    r_{bm} \coloneqq  -1.25 (v_z^{body})^2 - 0.4 |\omega_x^{body}| - 0.4 |\omega_y^B|.
    \end{equation}
We also penalized the angle between the $z$-axis of the world and the $z$-axis of the robot's body to maintain level body pose:
    \begin{equation}
        r_{ori} =  \arccos(R_b(3,3) )^2,
    \end{equation}
    where $R_b(3,3)$ is the last element of the rotation matrix representation of the body orientation.
We also motivated the policy to keep the height of the robot's base above the ground ($h_{base}$) around \unit[0.55]{m} with the tolerance of \unit[0.05]{m}:
    \begin{equation}
            r_{h} = max(0.0, \lvert h_{base} - 0.55 \rvert - 0.05).
    \end{equation}

\subsubsection*{ Regularization Rewards}

We used various regularization rewards.
We penalized the joint torques to prevent damaging joint actuators during deployment and to reduce energy consumption ($\tau \propto \text{electric current}$):
    \begin{equation}
          r_{\tau} \coloneqq - \textstyle \sum_{i\in joints} \lvert\lvert \tau_i \rvert\rvert^2.
    \end{equation}
we also penalized joint velocity and acceleration to avoid vibrations:
   \begin{equation}
        r_{s} = - c_k\sum_{i=1}^{12} (
        \dot{q_i}^2 + 0.01 \ddot{q_{i}}^2
        ),
    \end{equation}
    where $\dot{q_i}$ and $\ddot{q_{i}}$ are the joint velocity and acceleration, respectively.

    The magnitude of the first and second order finite difference derivatives of the target joint positions are penalized such that the generated joint trajectories become smoother:
     \begin{equation}
        r_{s} = - c_k\sum_{i=1}^{12} (
        (q_{i, t, des} - q_{i, t-1, des})^2 + 
        (q_{i, t, des} - 2q_{i, t-1, des} + q_{i, t-2, des})^2
        ),
    \end{equation}
    where $q_{i,t,des}$ is the joint target position of joint $i$ at time step $t$.

    We enforced soft position constraints in the joint space. To avoid the knee joint flipping in the opposite direction, we give a penalty for exceeding a threshold:
           \begin{eqnarray}
                r_{jc,i} &=& \begin{cases}
                -(q_i - q_{i, th})^2, & \text{if } q_i > q_{i, th}\\
                0.0 & \text{otherwise}
             \end{cases}, \\
                r_{jc} &=& \sum_{i=1}^{12} r_{jc,i},
            \end{eqnarray}
            where $q_{i, th}$ is a threshold value for the $i$th joint.
            We only set thresholds for the knee joint.
            
    Contacts with the environment were penalized except for the wheels:
    \begin{equation}
        r_{bc} \coloneqq -\lvert I_{c, body} \backslash I_{c, wheel} \rvert.
    \end{equation}

    Not terminating was densely rewarded:
    \begin{equation}
        r_{h, surv} \coloneqq 1.0 \quad \text{while not terminated}.
    \end{equation}

\subsection*{Comparison to Related Works and Validation of Our Method}

In this section, we present a survey of relevant research concerning the development of navigation policies for mobile robots using \ac{RL}.
Focused on the works based on model-free \ac{RL}, we extract essential design choices and conduct a comparative analysis with our approach.

It's noteworthy that alternative methodologies also exist, including those based on offline \ac{RL} or model-based \ac{RL}~\cite{kahn2021badgr,kahn2021land,kim2022learning}. For instance, Kahn et al.~\cite{kahn2021badgr, kahn2021land} showed outdoor navigation using an offline-trained dynamics model combined with a sampling-based planner. However, we maintain our focus on the model-free approach and local navigation setup (up to \unit[20]{m} distance to goal) because we aim to develop a highly responsive control policy with a high control rate and minimal onboard planning as described in the introduction.

\subsubsection*{Key Design Choices in Existing Literature}

Upon reviewing the existing literature, we have identified several critical factors that contribute to enhancing the navigation performance of a learned agent. 

Firstly, modularity and abstraction were stressed in many existing works.
Navigation problems are often addressed by decomposing them into sub-problems and then tackling each using specialized sub-modules. This modular approach, exemplified by hierarchical control systems, streamlines the developmental process as evidenced in existing studies~\cite{jain2019hierarchical,jain2020pixels,truong2021learning}.
In the context of \ac{HRL}, separating low-level locomotion and high-level navigation on different time scales benefits exploration and performance~\cite{jain2019hierarchical, nachum2019does}, with higher-level agents operating at a slower frequency. The temporal abstraction by design enhances exploration and improves final performance in some cases.
Furthermore, integrating pre-trained perception modules has shown to enhance the navigation proficiency in complex settings, as demonstrated in the investigations carried out by Müller et al.\cite{muller2018driving} and Hoeller et al.\cite{hoeller2021learning}. 
    
Our approach aligns with this paradigm. \ac{HLC} operates at a slower frequency and utilizes the pre-trained \ac{RNN} encoder from the \ac{LLC} as a state representation. Our experiment below shows the importance of the former in navigation performance, although the latter plays a crucial role in robust robot deployment, especially on challenging terrains (depicted in Move S4).

Secondly, many existing works utilize expert demonstrations, often sourced from either executing a sampling-based planner~\cite{pfeiffer2018reinforced} or human demonstrations~\cite{manderson2020vision, wang2023diffusion}.
    Imitation learning is a widely adopted approach for autonomous driving and navigation domains, offering accelerated learning and enhanced performance.
    
    In our case, even though we could generate expert demonstrations using a sampling-based planner~\cite{wellhausen2022ART}, we refrained from this approach due to its high computational overhead. Instead, we opted to leverage pre-generated paths from our simulation environment and trained a high-level policy to adhere to them via path sampling and dense reward.

Thirdly, memory, whether explicit or implicit, plays an important role in point-goal navigation. Literature frequently incorporates explicit memory mechanisms~\cite{savinov2018semi,fu2022coupling} or employs \ac{RNN} architectures to address this need~\cite{wijmans2019dd,choi2019deep,hoeller2021learning}. Extensive analysis by Wijmans et al.~\cite{wijmans2023emergence} highlights  the contribution of memory in successfully accomplishing navigation tasks.
    
    Similarly, we incorporate memory, albeit in a simplified manner. Tailoring our approach to real-world requirements, we designed our formulation to employ simple models (for example, \ac{MLP} or shallow \ac{CNN}) and interpretable states. Rather than depending on generic, navigation-agnostic \ac{RNN} structures, we integrate explicit information about visited positions and times.

Lastly,  dense reward-shaping is a prevalent strategy due to the inherent difficulty in training sparse reward formulations. 
    Notably, dense reward functions are frequently employed to incentivize policy progression or to penalize collision occurrences~\cite{zhu2021deep,surmann2020deep,jain2020pixels}.
    Some works define the dense reward functions based on the geodesic distance, which is the shortest obstacle-free path to the goal~\cite{wijmans2019dd,truong2021learning,pfeiffer2018reinforced}.
    
    In our approach, we leverage dense rewards only during the initial training phase. This is because the dense rewards based on the shortest obstacle-free path do not account for dynamic obstacles or intricate environmental dynamics such as varying friction and disturbances.

Though it's not mentioned above, similarly to locomotion research, the sim-to-real approach is widely adopted. Using large amounts of synthetic data improves training efficiency and robustness, owing to the large amount of data collected from diverse simulated scenarios.

We have integrated the above principles from the literature into our approach.
Our \ac{HRL} formulation follows the task decomposition appearing in the literature, and our simulation environment is designed based on the insights from points above.

\subsubsection*{Validation of Our Approach}

We proceed to conduct a comparative analysis between our approach and the baselines defined below. 
The first two baselines aim to validate the effectiveness of our graph-guided navigation learning approach, whereas the subsequent baselines are derived from existing literature. 
These subsequent baselines are meant to isolate and identify the distinct contributions of each component within our approach.

We first compared with a policy trained without path sampling. This baseline evaluates the advantage of using pre-generated, obstacle-free paths to generate waypoints during training. Even though it is trained in an environment identical to ours, the goals are uniformly distributed across the terrain without accounting for obstacles or ensuring a feasible path.

Then we compared ours to a policy trained in the environment without \ac{WFC} Features. This baseline evaluates the importance of providing various navigation challenges during the training. Instead of using terrain features generated by \ac{WFC}, this baseline is trained over rough terrains with randomly placed obstacles.

To evaluate the role of memory in our approach, we trained a memoryless baseline. This baseline does not incorporate position history, aligning it with the reactive policies presented by Jain et al.\cite{jain2020pixels} and Pfeiffer et al.\cite{pfeiffer2018reinforced}. It is trained in the same environment as ours.

We also evaluated the influence of the temporal abstraction to the navigagion performance. This baseline evaluates the importance of temporal abstraction for the navigation task. We train a high-level policy with the same control frequency as the low-level locomotion policy (\unit[50]{Hz}).

Lastly, an end-to-end baseline. This baseline assesses the advantages of the hierarchical decomposition of the task. It is a single policy directly outputting joint control commands, trained to pursue waypoints. The policy is trained with $r_h + r_l$, without the velocity tracking rewards. This formulation is in line with the work of Yang et al.\cite{yang2021learning} and Rudin et al.\cite{rudin2022advanced}.

\begin{table}
\centering

\resizebox{\columnwidth}{!}{%
\begin{tabular}{l|c|c}
\hline
\multirow{2}{*}{Policies}& \multicolumn{2}{c}{SPL (Success Rate) by Path Length} \\  \cline{2-3}
 &  5 to 10~\unit[]{m} &  10 to 20~\unit[]{m}\\
\hline
\textbf{Ours} & \textbf{0.897 (0.901)}& \textbf{0.689 (0.763)}\\ 
  
Baseline 1. No path sampling & 0.858 (0.840)& 0.497 (0.559)\\
Baseline 2. No WFC features & 0.865 (0.871)& 0.302 (0.305)\\

Baseline 3. Memoryless & 0.873 (0.897) & 0.526 (0.573)\\
Baseline 4. No temporal abstraction & 0.798 (0.823)& 0.370 (0.397)\\
Baseline 5. End-to-end & 0.304 (0.318)& 0.045 (0.046)\\

\hline
\end{tabular}
}
\caption{\textbf{Performance comparison between different navigation policies.} Values in the parenthesis indicate success rates. Each value represents an average taken from 1000 randomly generated terrains and paths.}
\label{tab:spl_comparison}
\end{table}

We conduct evaluations based on the methodology proposed by Anderson et al.~\cite{anderson2018evaluation}, employing the \acf{SPL}. The \ac{SPL} is given by:
\begin{equation}
    \frac{1}{N} \sum_{i = 1}^N S_i \frac{l_i}{\max(p_i, l_i)}
\end{equation}
where $l_i$ corresponds to the shortest path distance between the starting position and the goal in episode $i$, $p_i$ represents the actual path length traversed by the robot, and $S_i$ is a binary indicator denoting the success of episode $i$.
In our experiment, we compute $l_i$ as the shortest distance on the navigation graph.

We evaluated the \ac{SPL} and success rate for different path lengths ($l_i$), as shown in Table~\ref{tab:spl_comparison}. This was done using 1,000 randomly generated terrains, with randomly sampled paths ranging from 5 to 20~\unit[]{m}. A waypoint is given at the path's endpoint. An episode is considered successful when the robot approaches the waypoint within \unit[50]{cm} in \unit[60]{s}.

We begin by validating our graph-guided navigation learning approach with baselines 1 and 2.
Both baselines show degraded performance for the distant goals.
When a policy is trained to track arbitrary goals (baseline 1), we observed that the policy becomes overly conservative with distant goals.
This is due to the high occurrence of infeasible goals, leading to increased failures during training.
Without \ac{WFC}-generated terrain features during training (baseline 2), the randomly generated environment fails to provide a policy with diverse challenges like tight spaces and transitioning terrains. Consequently, baseline 2 exhibits limited performance when navigating complex environments.
Results from these baselines underscore the importance of a well-structured training environment that offers quality training data.

We then evaluate each component of our navigation policy. 
As given in Table~\ref{tab:spl_comparison}, our approach shows higher \ac{SPL} and success rates compared to the ablated baselines. 
When the memory is removed, the baseline 3 showed approximately \unit[16]{\%} lower \ac{SPL} for the distant goals (beyond \unit[10]{m}).  
The baseline 3 still shows more than \unit[50]{\%} success thanks to its explorative behavior, depicted in Fig.~\ref{fig:planner_comp}D-ii. 
The limited influence on performance can be attributed to our focus on local navigation. 
However, in scenarios that demand extensive memory capabilities, as demonstrated in \cite{wijmans2019dd}, this component becomes more important. 
Baselines 4 and 5 results show the importance of the hierarchical decomposition of the problem.
Baseline 4, lacking temporal abstraction, and baseline 5, omitting problem sub-division (modularity), both encounter increased difficulty in solving complex navigation tasks with distant goals.
Regarding baseline 5, the need to address rough-terrain locomotion and point-goal navigation within a single \ac{MDP} introduces challenges in terms of reward shaping. To achieve better results, extra engineering efforts would be necessary.

In conclusion, we have validated our approach across different training environments and \ac{MDP} formulations from existing literature.
The analysis shows that each individual component is important in enhancing navigation performance.
Our approach effectively incorporates key concepts from existing literature. In particular, our terrain generation and learning strategy substantially contribute to the final performance.

\subsection*{Comparison of Different Architectures}

\begin{figure}
    \centering
    \includegraphics[width=\columnwidth]{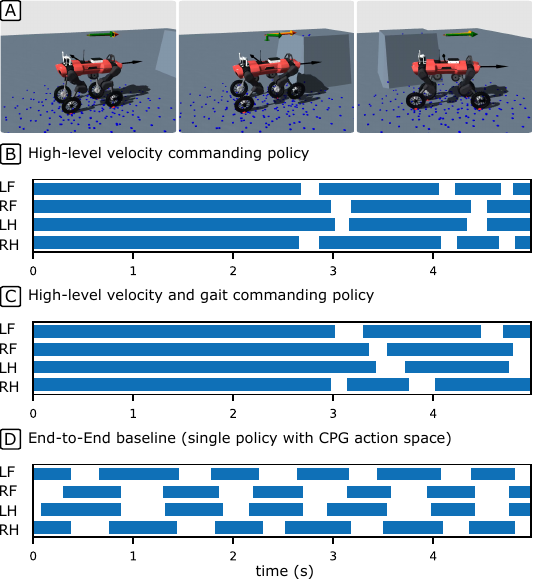}
    \caption{
    \textbf{Different controllers during collision avoidance}
    On flat terrain with an obstacle, the waypoint is given behind the obstacle. 
    \textbf{(A)} Motion sequence of our controller. 
    \textbf{(B-D)} Foot contact sequences of different approaches. The robot faces the obstacle at around \unit[2.0]{s}.}
    \label{fig:gaitswitch}
\end{figure}

We explored two other approaches to learning gait-switching behavior.

Firstly, we experimented with the hierarchical gait selection. In this design, High-level policy outputs desired foot contact states per foot (either 1 or 0) in addition to the velocity command. Low-level policy learns gait following and velocity tracking.  The low-level policy is first trained with known gait patterns like trot, pace, and static walk, and then trained alternatingly with the high-level policy.

    For the low-level policy, we introduced an additional gait-tracking reward, defined as 
\begin{equation}
        r_{gait} \coloneqq 0.1 \cdot \sum_{i\in{0,1,2,3}}
        \mathds{1}({fc(i)=fc(i)_{g}}),
    \end{equation} where $fc_(i)$ denotes the desired contact state of the $i$-th foot and  $fc_(i)_{g}$ is the target contact state given by the high-level policy.

Secondly, a \ac{CPG}-based baseline~\cite{lee2020learning, miki2022learning}. This is an end-to-end approach, with single policy. Trained with $r_h + r_r$. We used the same action space as Miki et al.~\cite{miki2022learning}. The gait frequency and duty factor are fixed to \unit[1.0]{Hz} and 0.5, respectively, and the initial phases are randomized.

We conducted an experiment where the robot is commanded toward a waypoint behind a $\unit[1]{m} \times \unit[1]{m}$ obstacle. The experimental setup and motion sequences are shown in Fig.\ref{fig:gaitswitch}.

Hierarchical controllers only stepped when faced with the obstacle (Fig.\ref{fig:gaitswitch}C and Fig.\ref{fig:gaitswitch}D).They exhibited similar behaviors with different gait frequencies and timing. 
On the other hand, the baseline showed regular stepping and continued stepping even when it is not necessary. This is mainly due to the fixed \ac{CPG} that limits the exploration of different walking patterns. In \cite{lee2020learning} and \cite{miki2022learning}, the gait frequency is manually set to 0 when no stepping is desired.

Our final design removed the gait selection for simplicity, but both resulted in similar performance and gait-switching behavior. 

The hierarchical gait selection approach follows the traditional separation of locomotion and gait planning~\cite{gangapurwala2022rloc, bellicoso2018dynamic, bjelonic2022planning}. The modular design simplifies the locomotion control problem with a fixed gait and allows for individual gait analysis~\cite{kolvenbach2018efficient}.  
To learn gait patterns, we used a learned action space that maps the output of the high-level policy to a distribution of gait parameters~\cite{ward2019improving,zhou2020plas, allshire2021laser}.
The generative model was trained with known gait parameters~\cite{bjelonic2019keep, bellicoso2018dynamic}.

\subsubsection*{Learned Action Space for Gait-generating High-level Policy}
\label{gait_generator}
For the gait commanding high-level policy, we had to implement a special action space.
Exploring the space of gait parameters with the commonly used Gaussian distribution can be inefficient because not all the real-valued vectors can represent feasible gaits, and the feasible parameters can be sparsely distributed.
To improve exploration and accelerate learning, we use a learned gait generator as the action space of the high-level policy.

Existing works have proposed using generative models such as \ac{VAE}s~\cite{zhou2020plas, allshire2021laser} or a normalizing flow~\cite{ward2019improving} to transform the action distribution into a different, possibly multi-modal, distribution.
Wenxuan et al.~\cite{zhou2020plas} and Allshire et al.~\cite{allshire2021laser} proposed to pre-train generative models with existing motion data for higher sample efficiency.

Similarly, we construct a learned latent action space with a RealNVP model~\cite{ward2019improving} that generates gait patterns from a Beta distribution. 
We chose RealNVP instead of \ac{VAE}~\cite{kingma2013auto} because the RealNVP can be updated during the \ac{RL} update by policy gradient thanks to its invertibility~\cite{dinh2016density, ward2019improving}.

We construct a stochastic policy $\pi(a|s)$ by two neural network modules in series. Firstly, an \ac{MLP} outputs parameters for the Beta distribution that serves as a base distribution.
Then follows an invertible normalizing flow layer to get $a = f_{\psi}(z)$, where $ z \sim \mathcal{N} (\mu_{\theta}(s), \sigma_{\theta}(s))$. $f_{\psi}$ denotes a RealNVP.
We can directly use the RealNVP policy instead of Gaussian policies within \ac{RL} algorithms since it is possible to compute the log-likelihood of the action by
\begin{equation}
    \log \big( \pi(a|s) \big)= \log \big(p_z(f_{\psi}^{-1}(a)) \big)
    + \log \left(\ \left| \det \left(\frac{\partial f_{\psi}^{-1}(a)}{ \partial a^T}\right) \right| \right).
\end{equation}

The RealNVP layers are pre-trained to generate gait parameters from a uniform distribution. It is trained by minimizing the log-likelihood:
\begin{equation}
    \mathbb{E}_x \left\{ - \log \big( p_z(f_{\psi}^{-1}(x)) \big)
    - \log \left(\ \left| \det \big({\partial f_{\psi}^{-1}(x)}/{ \partial x^T}\big) \right| \right) \right\} ,
\end{equation}
where $x$ is sampled uniformly from known gait parameters.

\subsection*{Filtering Terrain Parameters}

To ensure that the low-level policy training focuses on traversable terrain, we employ the adaptive terrain curriculum method introduced by Lee et al.~\cite{lee2020learning}. This selection process specifically targets the low-level policy training phase.

Using a genetic algorithm based on the \ac{MC}~\cite{brant2017minimal}, we avoid terrain parameters that are either too difficult or too easy for the agents. The fitness function, denoted as $f(c_\mathcal{T},\pi)$, is defined as follows:
\begin{equation}
f(c_\mathcal{T},\pi) = 
\begin{cases}
\mathbb{E} \{ \nu(s_t \mid c_\mathcal{T}) \}  & \text{if} \quad t_l < \mathbb{E}\{ \nu(s_t \mid c_\mathcal{T}) \}  < t_h\\
0.0 & \text{otherwise}
 \end{cases}.
\end{equation}

Here, $c_\mathcal{T}$ denotes the terrain parameter being evaluated, and $\pi$ represents the policy being trained. The expected value $\mathbb{E}{ \nu(s_t \mid c_\mathcal{T}) }$ is computed over the trajectories generated by the policy during each iteration, where $\nu(s_t \mid c_\mathcal{T})$ is a score function reflecting the successful traversal of a sampled terrain at state $s_t$. In our case, $\nu(s_t)$ is set to 1.0 if the velocity tracking error is less than 20\% of the command speed.

The threshold parameters $t_l$ and $t_h$ define the \ac{MC}, ensuring that terrain parameters with a success rate between $t_l$ and $t_h$ are selected. In other words, terrain parameters that fall within this success rate range are considered feasible for training.

These selected terrain parameters are then reused to generate tile maps for the subsequent high-level policy training, ensuring that the high-level policy is trained on feasible terrain environments suitable for navigation.

The concept of dynamic task generation and open-ended learning, demonstrated by the Open-Ended Learning Team at DeepMind~\cite{team2021open}, further supports the effectiveness of this approach. The automatic generation of new solvable problems enhances the agent's generalization capabilities.

\subsection*{Privileged Training }
We follow the privileged learning method proposed by Lee et al.~\cite{lee2020learning} for robust Sim-to-Real transfer.
The policy trained by \ac{RL} serve as "teacher policy". It uses the is the ground-truth state $s_t$ from simulation which includes privileged information $x_t$. $x_t$ includes ground friction coefficient or ground reaction forces, which are not directly observable in the real world.

A recurrent "student policy" network is trained in a supervised fashion without $x_t$. The student policy imitates the teacher and learns to construct an internal representation of the world from a sequence of the noisy real-world observations. 
A policy trained in this way has proven to be more adaptive and robust in real-world settings with high disturbances and noisy observations~\cite{lee2020learning, miki2022learning, loquercio2021learning}.

We employ the DAgger~\cite{ross2011reduction} algorithm for imitation learning.
We collect trajectories using the low-level student policy and label target actions using the teacher policy.
The loss function is defined as
\begin{equation}
    \mathcal{L} \coloneqq  \mathbb{E}_{(s_t, o_t)\sim\mathcal{D}} \left\{(\pi^{\text{teacher}}(s_t) - \pi^{\text{student}}(o_t, h_t))^2 \right\},    
\label{eq:student}\end{equation}
where $o_t$ denotes the observation and $h_t$ denotes the hidden state of the student policy. $o_t$ is a noisy version of $s_t \setminus x_t$.

\subsection*{Bounded Action Space}

We employ the Beta Distribution for the action space of \ac{HLC}~\cite{chou2017improving}. The beta distribution is a continuous probability distribution defined on the interval [0, 1], characterized by two shape parameters, $\alpha$ (alpha) and $\beta$ (beta). These parameters determine the shape and probabilities associated with different outcomes.

The \ac{PDF} of the beta distribution is given by the formula $f(x; \alpha, \beta) = x^{\alpha-1} (1-x)^{\beta-1}/ {B(\alpha, \beta)}$, where $x$ represents the random variable representing the probability, and $B(\alpha, \beta)$ is the beta function. The expected value (mean) of the beta distribution is given by $E[X] = {\alpha}/{(\alpha + \beta)}$, and the shape parameters $\alpha$ and $\beta$ determine the variance of the distribution. Higher values of $\alpha + \beta$ lead to lower variance.

We modify the parameterization of the beta distribution to define the bounded action space for our high-level policy. Instead of directly outputting the $\alpha$ and $\beta$ parameters from $\pi_{hi}$, we design our policy to directly output the mean and the sum of $\alpha$ and $\beta$. More specifically, let $a_1$ and $a_2$ be the outputs of the high-level policy $\pi_{hi}(s_t)$, we have $\alpha = a_1 \cdot a_2$ and $\beta = a_2 - (a_1 \cdot a_2)$.
This design choice eliminates the need for additional computing of the mean after inference.

\begin{table}
\centering
 \begin{tabular}{|c|c|}
\hline 
 Parameter & Value  \\ \hline
 discount factor & 0.99 \\
 KL-d target & 0.01  \\
 clip range & 0.2 \\
 entropy coefficient & 0.001 \\
max. episode length (s) & 10.0 \\
dt (s) & 0.02 \\
batch size & 500000 \\ 
num. minibatches & 20 \\ 
num. epochs & 4 \\ 
learning rate & adaptive$^*$ \\ \hline
\end{tabular}
\caption{\textbf{Hyperparameters for LLC teacher policy training.} ($*$) Follows the implementation by Rudin et al.~\cite{rudin2022learning}.}
\label{tab:hyperparametersteacher}
\end{table}

\begin{table}
\centering
 \begin{tabular}{|c|c|}
\hline 
 Parameter & Value  \\ \hline
 discount factor & 0.991 \\
 KL-d target & 0.01  \\
 clip range & 0.2 \\
 entropy coefficient & 0.001 \\
max. episode length (s) & 15.0 \\
dt (s) & 0.1 \\
batch size & 150000 \\ 
num. minibatches & 10 \\ 
num. epochs & 5 \\ 
learning rate & adaptive \\ \hline
\end{tabular}
\caption{\textbf{Hyperparameters for HLC training.}}
\label{tab:hyperparametersteacher2}
\end{table}
\end{document}